\documentclass[1p]{elsarticle}

\usepackage{lineno,hyperref}

\journal{ }

\usepackage{booktabs} 
\usepackage{amsthm}  
\usepackage{amsmath}
\usepackage{amsfonts}
\usepackage{amssymb}
\usepackage{bm}
\usepackage{comment}
\usepackage{algorithm}%
\usepackage{graphicx}
\usepackage{algpseudocode}
\usepackage{hyperref} 
\usepackage{xcolor}

\theoremstyle{remark}

\DeclareMathOperator{\Nodes}{Nodes}

\DeclareMathOperator{\Index}{Index}
\DeclareMathOperator{\Int}{Int}
\DeclareMathOperator{\Var}{Var}
\DeclareMathOperator{\Corners}{Corners}

\begin{document}

\begin{frontmatter}

\title{An Inextensible Model for Robotic Simulations of  Textiles}


\author[mymainaddress]{Franco Coltraro\corref{mycorrespondingauthor}}
\cortext[mycorrespondingauthor]{Corresponding author}
\ead{fcoltraro@iri.upc.edu}

\author[mysecondaryaddress]{Jaume Amor\'os}
\author[mymainaddress]{Maria Alberich-Carramiñana}
\author[mymainaddress]{Carme Torras}

\address[mymainaddress]{Institut de Rob\`otica i Inform\`atica Industrial, CSIC-UPC, Barcelona, Spain.}
\address[mysecondaryaddress]{Universitat Polit\`ecnica de Catalunya, Barcelona, Spain.}

\begin{abstract}
We introduce a new isometric strain model for the study of the dynamics of cloth garments in a moderate stress environment, such as robotic manipulation in the neighborhood of humans. This model treats textiles as surfaces which are inextensible, admitting only isometric motions. Inextensibility is imposed in a continuous setting, prior to any discretization, which gives consistency with respect to re-meshing and prevents the problem of locking even with coarse meshes. The simulations of robotic manipulation using the model are compared to the actual manipulation in the real world, finding that the error between the simulated and real position of each point in the garment is lower than 1cm in average, even when a coarse mesh is used. Aerodynamic contributions to motion are incorporated to the model through the virtual uncoupling of the inertial and gravitational mass of the garment. This approach results in an accurate, as compared to reality, description of cloth motion incorporating aerodynamic effects by using only two parameters.
\end{abstract}


\end{frontmatter}

\nolinenumbers

\section{Introduction}

Robotic manipulation of cloth in a domestic environment, i.e. not in its manufacturing plant, is an increasingly relevant problem because of the ubiquity and versatility of textiles in human lives and activities.

Nevertheless, textiles are difficult to model from a material viewpoint. They can be naturally treated as thin plates or shells, but they show an extreme dichotomy: 
they stretch as an elastic solid in the tangent directions, but bend almost freely in the out-of-plane ones. This combination means that textiles are very prone to buckling under the slightest compression. Moreover, they are usually non-isotropic because of their weaved inner structure, and nonlinear responses to stresses appear very soon \cite{HGB69}. An additional difficulty for the modeling of the motion of cloth in typical everyday uses is posed by aerodynamics: most textiles are light, so they are considerably affected by the resistance of air to their movement. One can try simple drag-and-lift models of this resistance in the surface of a cloth away from its edges \cite{Mao2009}, but those edges bend easily and thus interact with flow turbulences around them in a much more complicated way than a solid rigid, even a vibrating one \cite{Shelley:2011}.

\smallskip

The purpose of this work is to introduce and discuss a model of cloth intended for its control under robotic manipulation in a human environment, which means that the textiles will be subjected to moderate to low stresses \cite{Borras2020}. Because of its control purpose, the authors sought a model with which simulations of the motion of cloth can be computed faster than real time, with a margin of error of below 1 cm in the position of every point of the cloth when compared to its position in the real cloth subjected to the same manipulation by the robot.
Thus the main aims of this model are realism and speed, while dramatic effects and aesthetic considerations will be disregarded. 

\smallskip

We will model textiles as surfaces. To achieve computational speed taking advantage of the allowed cm margin of error led the authors to the most basic hypothesis in the model: we will assume that textiles are {\em inextensible}, that is, the surfaces that represent them only deform isometrically through space, preserving not only the area but also both dimensions of each piece of the cloth. This assumption simplifies the model by removing all considerations of elasticity in it, at a cost of introducing errors in the mm range in the position of the cloth under the low stresses at which domestic robots operate. Our constraints are derived from continuous conditions and hence the model is very stable under different resolutions of the meshed fabric, allowing its successful use in coarse-mesh simulations. We explain how to efficiently discretize the constraints using the Finite Element Method. The model keeps the relative change of area during simulations of the textiles under $1\%$ without exhibiting locking.    

\smallskip

Since the publication of the seminal paper \cite{Goldenthal:2007:ESI} a lot of attention has been given to the simulation of inextensible and quasi-inextensible textiles. The main difficulty of inextensible cloth simulation is \textit{locking}, i.e. the meshed garment becoming very rigid or facing a spurious resistance to bending. With limited exceptions such as creasing along straight lines, this is an unavoidable consequence of applying elasticity or spring models to a triangulated fabric and then tuning them for full inextensibility: in the inelastic limit, the lengths of all edges in the mesh become fixed, and often the only deformations preserving all these lengths are those in which the polyhedral surface moves as a rigid body. The underlying reason is the classical Cauchy Rigidity Theorem for polytopes, extended by Gluck \cite{Glu75}, who showed that generic closed polyhedral surfaces in space are rigid, i.e. they lock once you fix the length of their edges.\\

To overcome this difficulty, two approaches have been proposed in the literature: in \cite{Goldenthal:2007:ESI} a discrete model is presented where a quad-mesh discretized cloth is made inextensible in the horizontal and vertical directions but still elastic diagonally, i.e. allowing shearing but not stretching. For triangle meshes,  \cite{English:2008:ADS} use non-conforming triangles to impose full inextensibility, albeit sacrificing continuity of the polyhedral surface at the edges. One may wonder then, why is another inextensible model necessary? We give several reasons for this:

\begin{enumerate}
	
	\item To our knowledge, all inextensible or quasi-inextensible models in literature are discrete in nature, i.e. they impose inextensibility constraining areas, edge lengths, etc. This may create mesh related artifices \cite{Bender:2014:PBS} and often impedes convergence as the mesh is refined \cite{Nealen:2006:PDM}.

	\item The other family of models that are continuous, are all based on elasticity theory, with the drawback that, in the inelastic limit when Poisson's ratio tends to zero, exhibit locking \cite{Babuska:1992:LFE,Phillips2009:OLE}.

	\item As we already mentioned, the development of a continuous model that performs well with coarse meshes is useful, particularly in the context of robotic manipulation of deformable objects.
\end{enumerate}

\subsection{Contributions of this work}

\begin{description}
	
	\item[-] We present a continuous model for the simulation of inextensible textiles based on natural geometrical constraints. 
	
    \item[-] We derive in full detail a novel and efficient discretization of the model using the Finite Element Method (FEM). Our approach results in a unified treatment of triangles and quadrilaterals.
	
	\item[-] A framework for the  empirical validation of any cloth model using a robot and a simple way of accounting for the full aerodynamic contribution to dynamics in our cloth model.
	
\end{description}


\section{Related work}

The mechanic behavior of cloth has been extensively studied from the engineering and computer graphics viewpoints. The engineering focus has usually been on cloth manufacture and its mechanical resistance, using elasticity theory to examine local properties of the material. The aim in computer graphics has always been the simulation and representation of cloth motion, using mostly mass-spring models for speed of computation and paying attention to global problems such as nontrivial cloth topology and collisions. In both fields, the textiles are usually modeled as two-dimensional surfaces. Since the pioneering work of \cite{Terzopoulos:1987:EDM} dozens of different models have been proposed for cloth simulation. Our approach retakes the original idea of that paper: understanding cloth's internal dynamics as the preservation of the first fundamental form of the surface. For surveys and research problems see \cite{Choi:2005:RPC,Nealen:2006:PDM,Bender:2014:PBS}. We now review some of the most popular methods for simulating the internal dynamics of cloth, noting that there is not a clear cut separation between some of them.  We deliberately only reference lines of research which model internal dynamics of cloth in an essentially different \textit{physical} manner. Approaches that develop novel numerical methods to integrate existing physical models are not considered, since that is not the contribution of our paper.

\begin{description}
	
	\item[Textile engineering:] The modeling of fabrics from a Structural Mechanics viewpoint has been a topic of interest in Textile Engineering since the start of industrial manufacture of cloth. See \cite{HGB69} for a survey, and completion in many respects, of this theory, or \cite{Hu04} for updates. This modeling has been hampered by the complexity of fabric as a material: it is inhomogeneous, even discontinuous, already at a relatively sizable scale; highly anisotropic and capable of a nonlinear answer even to relatively modest strains; prone to buckling under the slightest compression. 
	
	Because of these complexities, investigations of the explicit mathematical expression of the stress-strain relationships of fabrics start from the elasticity theory of solids, and typically consider fabric as an elastic thin plate, more rarely as a shell. Ignoring its thickness, the resulting surface has 6 strain fields that determine its displacement and deformation, and a $6 \times 6$ symmetric stiffness matrix giving the constitutive equations that relate stress and strain (\cite{Hu04} p.10). Particular properties of fabric such as orthotropicity reduce the number of stiffness parameters to take into account and lead to simplifications of this general elastic model \cite{HGB69,WAY03}, or to the introduction of nonlinear responses in the model \cite{Lloyd80}. 
	
	The development of these models has been based foremost on experimentation, using procedures such as the Kawabata Evaluation System (KES) \cite{K82}, which accurately measures the tensile, shear, pure bending, compression and friction behavior of a cloth sample under stresses that are typical in the cloth manufacturing process,
	in order to establish the stiffness parameters of the model. The increase in available computing power has made viable, although not for real time computing, sophistications of this basic model to take into account the fiber structure of yarns and fabrics \cite{Otaduy:2014:YSC,Jiang:2017:AEC}.

	The computational complexity of these models opened the way to the development of simpler, descriptive models that would be later become of common use in Graphic Computer Science, treating  fabric as a viscoelastic material that can be modeled with spring-and-mass systems \cite{Paipetis81}. While such descriptive models have been extensively used in fabric simulation, their limited connection to reality has made them of limited use for the industrial handling of cloth. Here the thin plate with linear elasticity models prevail. 
	
	\item[Elasticity theory:] these models, e.g. \cite{Etzmuss:2003:FECM,Susin:2006:GPU,Wang:2011:DDE}, derive the internal forces of the garment from elasticity theory. They are all continuous in nature and have the advantage of being physically based (i.e. they use the stress and strain tensors), stable under different meshing of the garment and convergent when the mesh is refined \cite{Choi:2005:RPC}. Mostly, they use finite elements to discretize the equations of motion \cite{Braess:2007:FE}. They can be expensive to evaluate (especially when using non-linear elasticity or the co-rotational method, necessary for ensuring rotational invariance \cite{Susin:2006:GPU}) and may exhibit locking of triangular meshes if elasticity is heavily reduced. See \cite{English:2008:ADS,Jin:2017:IC} and more in general \cite{Babuska:1992:LFE,Phillips2009:OLE} for a more detailed discussion.
	
	\medskip
	
	\item[Mass-spring systems:] these models, e.g. \cite{Baraff:1998:LSC,Provot:2001:MSC,Choi:2002:SBR} derive internal forces from spring-like energies. They are cheap to evaluate and very intuitive, but less physically sound: they are mesh dependent, have a lot of (non-physical) parameters to be tuned (see \cite{Mongus:2012:HEA} for an evolutionary algorithm to tune them) and do not show a convergent behavior when the mesh is refined \cite{Nealen:2006:PDM}. 
	
	\medskip
	
	\item[Constrained dynamics:] these models derive internal forces from explicit conditions that the cloth must satisfy \cite{Bender:2014:PBS}. Most of the methods use some kind of Lagrange multipliers to impose the constraints. They mostly differ in what the conditions are and the algorithm used to impose the constraints. Our method fits in this category. Sometimes they are used as velocity filters that complement the previous methods. There are mainly two kinds: 
	
	\smallskip
	
	\begin{enumerate}
		
		\item \textit{Continuous}: in this case the constraints are continuous in nature, e.g. imposing bounds to the strain tensor. Examples of this approach are \cite{Thomaszewski:2009:CSL,Narain:2012:AAR,Guanghui:2015:ASL,Wang:2016:RSL}. To our knowledge all methods use elasticity theory to some extend, and therefore in the limit, when elasticity is reduced, face the same problems commented previously. Our method lies in this category with the important difference that it does not use elasticity, but differential geometry of the surface to derive the constraints. 
		
		\smallskip
		
		\item \textit{Discrete}: in this case constraints are discrete in nature, e.g. preserving the length of the edges of the meshed garment or the area of the elements. They are derived concretely for the mesh at hand. Examples are \cite{Goldenthal:2007:ESI,English:2008:ADS,Han:2015:APSL,Jin:2017:IC}. Their main drawback is the lack of convergence of the methods (as opposed to the continuous case) and the possible introduction of mesh related artifacts \cite{Guanghui:2015:ASL}. Nevertheless they can be fast and handle better the inelastic scenario than the elasticity-based continuous models.
		
	\end{enumerate}
	
	\medskip
	
	\item[Others:] There are two different recent trends: modeling woven cloth at the fiber level \cite{Otaduy:2014:YSC} as opposed to the macroscopic level (i.e. as a surface) and considering cloth as a non-Newtonian fluid using the Material Point Method (MPM) adapted to co-dimensional elasticity \cite{Jiang:2017:AEC}. The first method is computationally very intensive and the second while being more efficient depicts cloth as very elastic.
	
\end{description}

\smallskip

The definitive test for a model of cloth is its comparison to reality. Textile engineering models are focused on such comparison, to the point of developing specialized testing equipment. But the object of their study are local properties of cloth, such as elasticity parameters, which are tested in static scenarios \cite{Hu04,Miguel2012,Wang:2011:DDE,Clyde2017}. To the knowledge of the authors, none of the models developed in the
area of Computer Graphics for the simulation of the motion of textiles has been able to compare its results with the motion of cloth dynamically. Being oriented to robotic manipulation of cloth, our model is ideally placed for such experimental validation as discussed in Section \ref{s:expvalidation}.


\section{Overview}

\begin{figure}[htb]
	\centering
	\includegraphics[scale=.6]{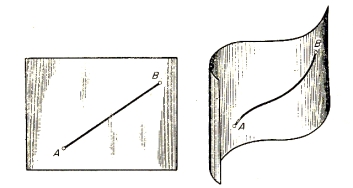}
	\caption{\label{fig:isometry}
		Isometric deformation of a surface: in both surfaces, the length of the depicted (geodesic) curve is the same. Image taken from \url{https://solitaryroad.com/c334.html}.}
\end{figure}

Our approach is to consider cloth as a \textit{continuous} and \textit{inextensible} two-dimensional surface. Formulating the model at the continuous level is important because it avoids as much as possible mesh dependencies: i.e. without changing the physical parameters of the model (e.g. stiffness, damping, etc.) the results are mostly independent of mesh topology (this will be checked in our experiments). This is not the case when using mass-spring systems \cite{Nealen:2006:PDM}. On the other hand, inextensibility is a very reasonable approximation for most textiles (specially in a robotics context, where fine details such as wrinkles are not needed). We will show this comparing our model to real-life experimental data. Moreover, this assumption reduces greatly the amount of tuneable parameters of the model (stretching and shearing \cite{HGB69,Baraff:1998:LSC} are no longer present). For us inextensibility means that over time the metric of the surface is preserved. This implies, that for all times the length of any given curve inside the surface will remain constant (see Figure \ref{fig:isometry}). Given a parametrization $\varphi=\varphi(\xi,\eta)\in\mathbb{R}^3$ of a smooth surface $S\subset\mathbb{R}^3$ its first fundamental form is given by:
\begin{equation*}
d\varphi^\intercal\cdot d\varphi = \begin{bmatrix}
\langle\varphi_{\xi},\varphi_{\xi}\rangle & \langle\varphi_{\xi},\varphi_{\eta}\rangle  \\
\langle\varphi_{\eta},\varphi_{\xi}\rangle  & \langle\varphi_{\eta},\varphi_{\eta}\rangle 
\end{bmatrix} = \begin{bmatrix}
E & F \\
F & G
\end{bmatrix},
\end{equation*}
where $\varphi_{\xi} = \partial_{\xi}\varphi$ denotes partial differentiation. It is well known that this $2\times2$ matrix defines the \textit{metric} of $S$, i.e., it gives a unique way for measuring angles and distances intrinsically in the surface \cite{do1976differential}.

\subsection{Isometric motion}

Assume now that $S$ is moving \textit{isometrically} through space; this means we have a continuum of surfaces $\{S_t\}_{t\geq0}$ such that $S_0 = S$ and for every $t>0$ there exists a (smooth) isometry $H_t:S_0\rightarrow S_{t}$. We think of the parameter $t\geq0$ as time. If $\varphi$ is any parametrization of $S$, then $\varphi(t) = H_t\circ\varphi$ is a parametrization of $S_t$ and isometry means:
\begin{equation}\label{inext_pos}
d\varphi(t)^\intercal\cdot d\varphi (t) = d\varphi^\intercal\cdot d\varphi \quad \text{for all}\quad t\geq 0.
\end{equation}
These are 3 independent conditions (depending on time and space):
\begin{equation}\label{constraints}
\langle\varphi_{\xi},\varphi_{\xi}\rangle(t) = E_0,\; \langle\varphi_{\xi},\varphi_{\eta}\rangle(t) = F_0,\; \langle\varphi_{\eta},\varphi_{\eta}\rangle(t) = G_0\; \text{for }t\geq 0.
\end{equation}
Moreover, noting that $d\varphi(t) = \nabla H_t|_{\varphi}\cdot d\varphi$, we can rewrite (\ref{inext_pos}) as 
\begin{equation}\label{isom_def}
d\varphi^\intercal\left(\nabla H_t|_{\varphi}^\intercal\cdot \nabla H_t|_{\varphi} - I\right)\cdot d\varphi = 0
\end{equation}
where $\nabla H_t$ is the Jacobian of $H_t$ and $I$ is the identity $3\times3$ matrix. Therefore if $H_t(x) = Ax + b$ is any rigid transformation of $\mathbb{R}^3$, we must have that $A$ is an orthogonal matrix and hence it preserves the metric of $S$ as expected. This rigidity invariance is not obtained when using linear elasticity models, e.g. big rotations can lead to different elastic restoring forces \cite{Susin:2006:GPU}. Also, regarding the relationship of Equation (\ref{isom_def}) with Elasticity Theory, particularly the \textit{strain tensor} (i.e. Cauchy-Green strain tensor minus the identity: $E = \tfrac{1}{2}\left[\nabla H_t|_{\varphi}^\intercal\cdot \nabla H_t|_{\varphi} - I\right]$), note that all deformations that have zero strain are isometric, but not all isometries are strain-free (e.g. bending along a line). This means that imposing equation (\ref{isom_def}) as a hard constraint (as we will do) is not equivalent to imposing zero strain deformations with classical elasticity theory.

\subsection{Cloth model:}
The rest of this paper is organized as follows: we explain how to impose conditions (\ref{constraints}) as hard constraints using \textit{Lagrange multipliers}. In order to treat computationally this, we employ finite elements \cite{Braess:2007:FE} in the following way: we use them to discretize the Lagrangian of the mechanical system but not the equations of motion, obtaining this way as \textit{Euler-Lagrange equations} an  ordinary differential equation (ODE) system and not a (difficult to integrate, e.g. \cite{Terzopoulos:1987:EDM}) partial differential equation (PDE) system. We then discuss the computational complexity of evaluating the inextensibility constraints. These will be the most technical parts of the manuscript and it is due to the proposed algorithm having aspects that are delicate and original from a mathematical viewpoint. Particularly important is how to constrain nodes lying at the boundaries (if not done properly this could lead to \textit{locking}) and the parametrization of the surface (this allows arbitrary topologies to be used). After this is done, we discuss how to integrate numerically the equations of motion, modifying the approach described in \cite{Goldenthal:2007:ESI} to consider at the same time inextensibility and collisions against a fixed object. Bending is modeled using Willmore's Energy as described in \cite{Bergou:2006:QBM}.

\subsection{Performance evaluation:}

We test the performance of the presented inextensible model in several challenging scenarios: first, we perform a simple quasi-static test to show the locking-free nature of our model, along with its independence  with respect to different meshed topologies. We use different triangle and quadrilateral meshings to prove our point. Second, we show how to simulate Cusick's test with our model. With this experiment we show the stability of our model when the mesh is refined. Lastly, we present a scenario with non-trivial cloth topology, where we simulate the motion of a tank-top and check to what extend our theoretical inextensibility assumptions are being met in practice, computing several area errors. 

\begin{figure}[!h]
	\centering
	\includegraphics[width=0.8\linewidth]{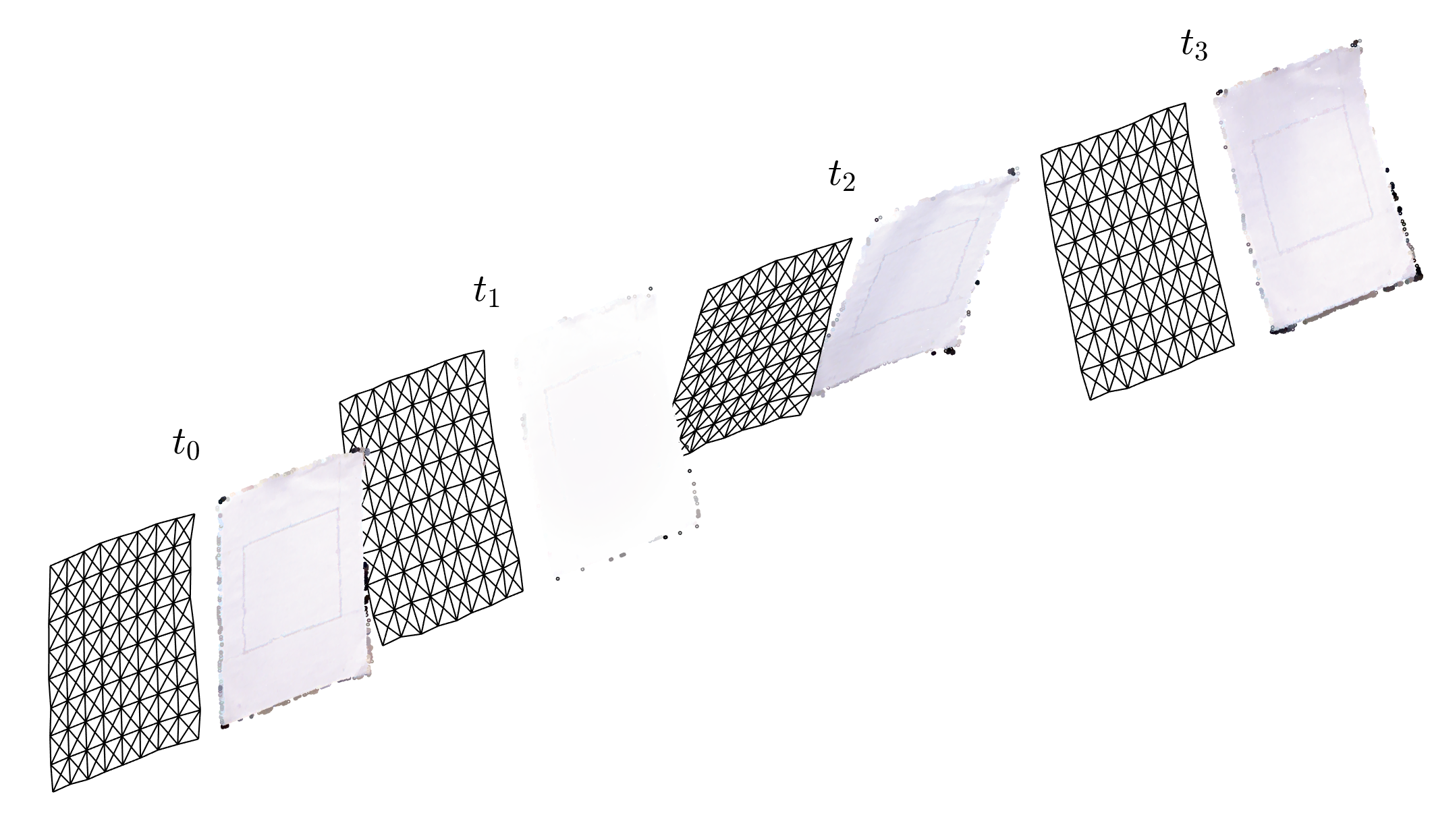}
	\caption{\label{paper}Comparison at four time instants of the recorded fast motion of paper (right) versus its simulation with the inextensible model (left) with $\delta = 0.37$ and $\alpha = 2.50$. The mean absolute error is $0.41$ cm.}
\end{figure} 

\subsection{Experimental validation}

\begin{figure}[htb]
	\centering
	\includegraphics[width=0.85\linewidth]{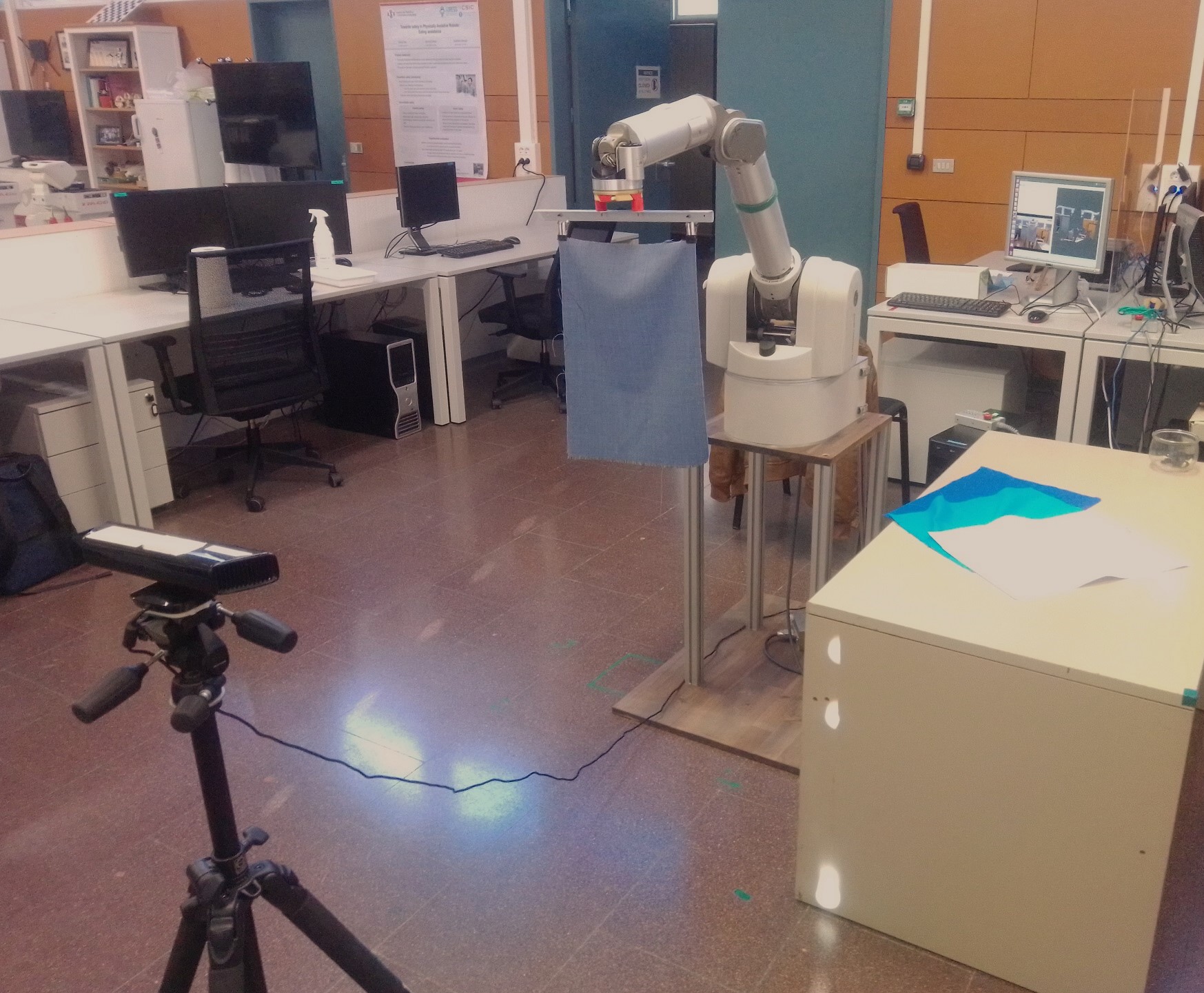}
	\caption{\label{setup}Experimental setup for the recording of real world data. On the left, the depth camera. On the right, the robotic arm with the affixed hanger.}
\end{figure}

Finally, we compare the inextensible model with reality (see Figure \ref{paper}) and show how with as few as two parameters, we are able to model different types of textiles (such as cotton, wool and felt) under fast and slow motions. This is the main content of the companion work \cite{Iros2021}. To perform this real-world validation we use a Barret robotic arm together with a depth camera to record the real motion of garments being shaken at different velocities (see Figure \ref{setup}).

We implement oscillatory motions: a forward and backwards shaking (see Figure \ref{paper}) at two different speeds. Afterwards, the resulting point-cloud recordings are de-noised, interpolated and meshed, so that we end up with the spatial trajectory of the vertices of a polyhedron. In order to keep the validation manageable, we focus on rectangular pieces of cloth for four different textiles: light cotton, wool, felt and paper. They are all of the same dimensions: $29.7\times42$ cm (DIN A3). We use a trivial rectangular topology in order to avoid occlusions in the point-cloud when using the depth camera to record the motions. Finally, we find the physical parameters of the model that best approximate the motion of the polyhedron using a least squares approximation and a minimization algorithm. We study the evolution of absolute errors and dispersion measures (i.e. standard deviations) of the real garments versus the simulations for all textiles and motions.

\section{Discretization of the surface}\label{disc_surf}
For the rest of the paper we will assume that our surface $S$ has been discretized, i.e., that we have a polyhedron consisting on a ensemble of  \textit{quadrilaterals or triangles} $\Omega_{e}$ for $e\in\{1,\ldots,n_q\}$ that approximates our original surface: $S \simeq \cup_e \Omega_{e}$ (see Figure \ref{fig:param_shorts}). For the sake of clarity we will assume the use of quadrilaterals, although everything works the same with triangles. Each element will have its own local parametrization:
\begin{figure}[htb]
	\centering
	\includegraphics[width=.4\linewidth]{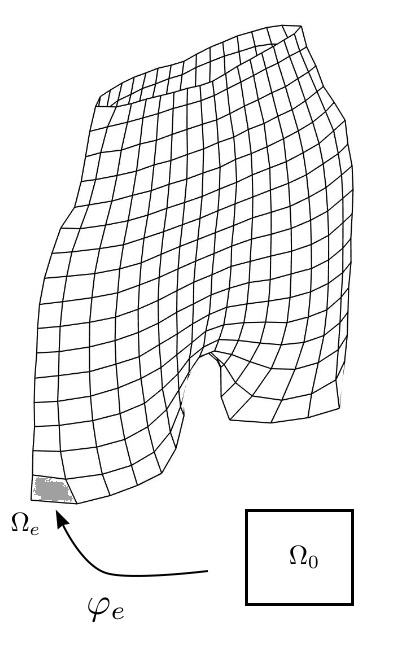}
	\caption{\label{fig:param_shorts} Bilinear parametrization of one of the elements of a quadrangulated surface (shorts).}
\end{figure}
\begin{equation}\label{quad_param}
\varphi_e:\Omega_0 := [-1,1]\times[-1,1]\rightarrow\Omega_{e};\quad \varphi_e(\xi,\eta) = \sum_{i=1}^{4}p_{i}^e N_{i}(\xi,\eta)
\end{equation} 
where the $p_{i}^e\in \Omega_{e}\subset\mathbb{R}^3$ are the 4 corners of the quadrilateral and the functions $N_{i}$ are called the shape functions:
\begin{align*}
N_{1} &= \frac{1}{4}(1-\xi)(1-\eta),\quad 
N_{2} = \frac{1}{4}(1+\xi)(1-\eta),\quad \\
N_{3} &= \frac{1}{4}(1+\xi)(1+\eta),\quad 
N_{4} = \frac{1}{4}(1-\xi)(1+\eta)
\end{align*}
and have the property that $N_{i}(q_j) = \delta_{ij}$ where $\delta_{ij}$ is Kronecker delta and the $q_j$'s are the four corners of $\Omega_0$ starting at $(-1,-1)$ and going counter clockwise. Hence, $\varphi_e(q_i) = p_{i}^e$. If the surface is moving through space, the parametrizations $\varphi_e = \varphi_e(t)$ depend of time and this means that $p_i^e = p_i^e(t)$ vary with time. Lets fix a \textit{global} ordering of the $n$ vertices of $S$. We will call them the nodes of the surface: $\Nodes(S) = [p_1,\dots,p_n]$, where $p_i = (x_i,y_i,z_i)\in\mathbb{R}^3$. Then, we can construct the vector of positions of the textile for any time $t\geq 0$:

\begin{align*}
\bm{\varphi}(t) &:= (\textbf{x}(t),\textbf{y}(t),\textbf{z}(t))^\intercal \\&= (x_1(t),\ldots,x_n(t)|y_1(t),\ldots,y_n(t)|z_1(t),\ldots,z_n(t))^\intercal\in\mathbb{R}^{3n}.
\end{align*}
\subsection{Indicator functions}

Now, we can define global (but non-smooth) indicator functions $\mathcal{N}_i:\cup_e \Omega_{e}\rightarrow\mathbb{R}$ such that $\mathcal{N}_i(p_j) = \delta_{ij}$ and given any element $\Omega_e\ni p_i$ we have that $\mathcal{N}_i|_{\Omega_e} = N_k$ where $N_k(\varphi_e^{-1}(p_i)) = 1$. Thus, if $p = \varphi(t)\in S_t$ is any point of the moving surface, it can be written as
\begin{equation}\label{perorata}
\varphi(t) = \sum_{i=1}^{n}p_{i}(t) \cdot\mathcal{N}_{i}(p),
\end{equation}
and therefore $\partial_{\xi}\varphi(t) = \sum_{i=1}^{n}p_{i}(t) \partial_{\xi}\mathcal{N}_{i}\text{ and } \partial_{\eta}\varphi(t) = \sum_{i=1}^{n}p_{i}(t) \partial_{\eta}\mathcal{N}_{i}$. 
\medskip

To finish this section, we want to remark that usually with FEM methodologies one parametrizes a region $R\subset\mathbb{R}^2$ where the values of a function $\varphi$ are searched; however we parametrize each quadrilateral $\Omega_e\subset S$ independently. Therefore, with this approach $S$ can have any given topology. Also, since $\varphi_e(0,0) = \tfrac{1}{4}\sum_{i=1}^{4}p_{i}^e$, the mean of the nodes of each quadrilateral belongs to the element (even though the quadrilaterals are \textbf{not} flat, they are bi-linear), so when rendering we can get higher resolution by adding the middle point.

\section{Constraint function}\label{constr_sec}
We will construct a smooth and easy to evaluate constraint function
\begin{equation}\label{cons_func}
\textbf{C}:\Nodes(S_t)\simeq\mathbb{R}^{3n}\rightarrow\mathbb{R}^{n_c}
\end{equation}
such that it is identically zero when all the $n_c$ conditions ensuring inextensibility are satisfied. This function must depend only on the position vector of the nodes $\bm{\varphi}(t)$, and on nothing else. As an example, in the case of the isometric motion of a curve, we would preserve the lengths of all its edges, so for each edge $k$ we write 
\begin{equation}\label{modelo_grinspun}
\hat{C}_k(\bm{\varphi}(t)) = ||p_{k_1}(t) - p_{k_2}(t)||^2 - l_{k}^2,
\end{equation}
where $p_{k_1}, p_{k_2}$ are the two endpoints of edge $k$ and $l_{k}>0$ is the length of the edge at rest. We will apply these constraints to the boundary curves of our cloth $S$.

\subsection{Weighted Galerkin residual method}

On the other hand, we now explain how to discretize equation (\ref{constraints}) for constraining interior nodes: substituting equation (\ref{perorata}) in the left hand side of the first equation of (\ref{constraints}) we get:
\begin{equation*}
\langle\varphi_{\xi},\varphi_{\xi}\rangle(t) = \sum_{i,j=1}^{n}\langle p_{i}(t),p_{j}(t)\rangle\cdot\partial_{\xi}\mathcal{N}_{i}\partial_{\xi}\mathcal{N}_{j}.
\end{equation*}

The only problem with the previous equation is that it cannot be evaluated at the nodes because the derivative of the indicator functions is not defined there. Nevertheless, if we write:
\begin{equation*}
\langle\varphi_{\xi},\varphi_{\xi}\rangle(t) = \sum_{l=1}^{n}E_l(t)\cdot\mathcal{N}_{l},
\end{equation*}
where $E_l(t)$ are the values of the first coefficient of the metric at the nodes, and then multiply by any indicator function $\mathcal{N}_{k}$ and integrate over the surface, we obtain:
\begin{equation}\label{weight_galerkin0}
\sum_{l=1}^{n}E_l(t)\int_{S}\mathcal{N}_{k}\mathcal{N}_{l}dA =  \sum_{i,j=1}^{n}\langle p_{i}(t),p_{j}(t)\rangle\cdot\int_{S}\mathcal{N}_{k}\partial_{\xi}\mathcal{N}_{i}\partial_{\xi}\mathcal{N}_{j}dA,
\end{equation}
and therefore the coefficients $E_l(t)$ can be found by pre-multiplying the right hand side of (\ref{weight_galerkin0}) by the inverse of the \textit{mass matrix} 
\begin{equation}\label{mass_matrix}
M_{ij} = \int_{\cup_e\Omega_e}\mathcal{N}_i\mathcal{N}_j dA.
\end{equation} 

In practical implementations the mass matrix $M$ is substituted by a diagonal matrix called the \textit{lumped mass matrix} $M_L$ defined as 
\begin{equation}
(M_L)_{ii} = \sum_j M_{ij}.
\end{equation}

We can imagine this process as collapsing all the mass of the surface to the nodes \cite{Braess:2007:FE}. This is very convenient because it allows to compute the inverse of the mass matrix as the multiplication by the reciprocals of the constants $m_k := (M_L)_{kk}$. 

\begin{figure}[htb]
	\centering
	\includegraphics[width=0.9\linewidth]{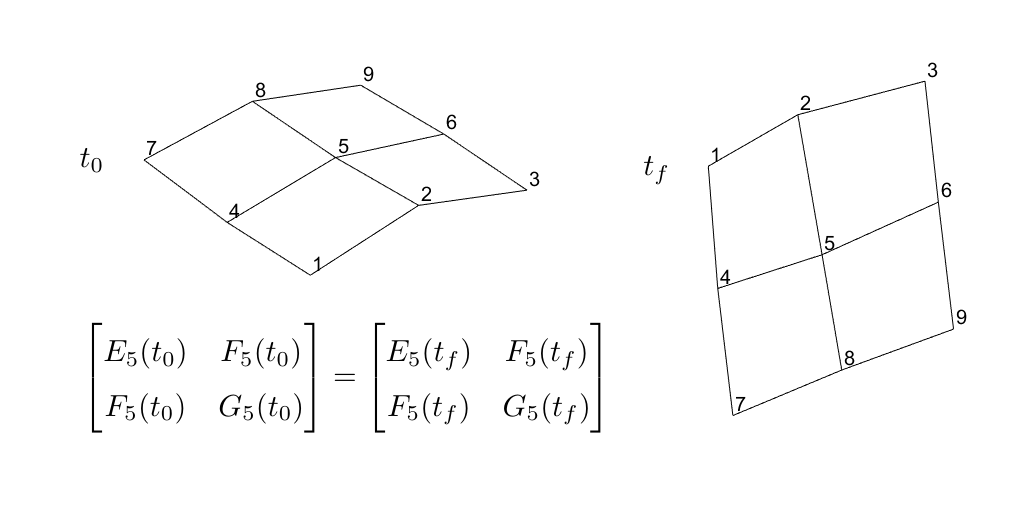}
	\caption{\label{fig:forma_fundamental}
		In order to have an isometry between the times $t_0$ and $t_f$, we impose the preservation of the first fundamental form at the central node $5$, we constraint the length of the boundary edges $\vec{12},\vec{23},\dots,\vec{74},\vec{41}$ and we enforce $F_k(t_0) = F_k(t_f)$ at the corners $k\in\{1,3,7,9\}$.}
\end{figure}

Therefore
\begin{equation}\label{weight_galerkin}
E_k(t) =  \frac{1}{m_k}\sum_{i,j=1}^{n}\langle p_{i}(t),p_{j}(t)\rangle\cdot\int_{S}\mathcal{N}_{k}\partial_{\xi}\mathcal{N}_{i}\partial_{\xi}\mathcal{N}_{j}dA.
\end{equation}

Naturally, similar equations are obtained when discretizing $\langle\varphi_{\eta},\varphi_{\eta}\rangle(t)$ and $\langle\varphi_{\xi},\varphi_{\eta}\rangle(t)$:

\begin{align*}
G_k(t) &=  \frac{1}{m_k}\sum_{i,j=1}^{n}\langle p_{i}(t),p_{j}(t)\rangle\cdot\int_{S}\mathcal{N}_{k}\partial_{\eta}\mathcal{N}_{i}\partial_{\eta}\mathcal{N}_{j}dA,\\
F_k(t) &=  \frac{1}{m_k}\sum_{i,j=1}^{n}\langle p_{i}(t),p_{j}(t)\rangle\cdot\int_{S}\mathcal{N}_{k}\partial_{\xi}\mathcal{N}_{i}\partial_{\eta}\mathcal{N}_{j}dA.
\end{align*}

In summary, to impose constraints (\ref{constraints}), for every interior node $k$ and $t>0$, we enforce $E_k(t) = E_k(0),\; F_k(t) = F_k(0),\; G_k(t) = G_k(0)$ (see Figure \ref{fig:forma_fundamental}).

\begin{algorithm}
	\begin{algorithmic}[1]
		\State {$m \gets \text{\#nodes of each element}$}
		\Comment(\textit{m = 3 (triangles) or m = 4 (quads)})
		\For {element $\Omega_e$} 
		\State {$\varphi_e := \sum_{i=1}^{4}p_{i}^e N_{i}$} \Comment(\textit{parametrization of }$\Omega_e$)
		\State {$I \gets \Index(\Omega_e)$}
		\Comment(\textit{node's indices of }$\Omega_e$ \textit{e.g.} $[4,48,50,13]$)
		\For {Gauss point $x_l$ and weight $w_l$}
		\State $E_l \gets \varphi_\xi(x_l)\cdot\varphi_\xi(x_l)$;
		\State $F_l \gets \varphi_\xi(x_l)\cdot\varphi_\eta(x_l)$; 
		\State $G_l \gets \varphi_\eta(x_l)\cdot\varphi_\eta(x_l)$;
		\State $dA_l \gets \sqrt{|E_l\cdot G_l - F_l^2|}\cdot w_l$;
		\For {$k,i,j=1,\dots,m$}
		\State $t^{kij}_F += N_k(x_l)\cdot\partial_\xi N_i(x_l)\cdot\partial_\xi N_j(x_l)\cdot dA_l$;
		\State $t^{kij}_F += \tfrac{1}{2}N_k(x_l)\cdot\left[\partial_\xi N_i(x_l)\cdot\partial_\eta N_j(x_l) + \partial_\eta N_i(x_l)\cdot\partial_\xi N_j(x_l)\right]\cdot dA_l$;
		\State $t^{kij}_G += N_k(x_l)\cdot\partial_\eta N_i(x_l)\cdot\partial_\eta N_j(x_l)\cdot dA_l$;                
		\EndFor
		\EndFor
		\For {$k,i,j=1,\dots,m$} \Comment(\textit{global assembly of the tensors})
		\State $\textbf{T}^{I(k)I(i)I(j)}_E +=  \frac{1}{m_{I(k)}}t^{kij}_E$; 
		\State $\textbf{T}^{I(k)I(i)I(j)}_F +=  \frac{1}{m_{I(k)}}t^{kij}_F$;
		\State $\textbf{T}^{I(k)I(i)I(j)}_G +=  \frac{1}{m_{I(k)}}t^{kij}_G$;
		\EndFor
		\EndFor
		\caption{Computation of $\textbf{T}_{E}$,$\textbf{T}_{F}$,$\textbf{T}_{G}$}\label{algo_tensor}
	\end{algorithmic}
\end{algorithm}

\subsection{Tensor definition}

We now discuss how to compute the right hand side of (\ref{weight_galerkin}). Let $\Int(S)$ denote the interior (non-boundary) nodes of $S$. We can define three \textit{time independent} 3-tensors that will allow us to compute the $3$ coefficients of the metric of $S$. First we have $\textbf{T}_E$:
\begin{equation*}
T^{kij}_E = \frac{1}{m_k}\int_{\cup_e\Omega_e}\mathcal{N}_k\partial_{\xi}\mathcal{N}_i\partial_{\xi}\mathcal{N}_j dA,
\end{equation*}
where $k\in\Int(S)$ and $i,j\in\Nodes(S)$. Similarly, $\textbf{T}_G$ is 
\begin{equation*}
T^{kij}_G = \frac{1}{m_k}\int_{\cup_e\Omega_e}\mathcal{N}_k\partial_{\eta}\mathcal{N}_i\partial_{\eta}\mathcal{N}_j dA,
\end{equation*}
and lastly $\textbf{T}_F$, where $k\in\Int(S)\cup\Corners(S)$, is
\begin{equation*}\label{Ftensor}
T^{kij}_F = \tfrac{1}{2 m_k}\int_{\cup_e\Omega_e}\mathcal{N}_k(\partial_{\xi}\mathcal{N}_i\partial_{\eta}\mathcal{N}_j + \partial_{\eta}\mathcal{N}_i\partial_{\xi}\mathcal{N}_j)dA.
\end{equation*}

This choice of the $\textbf{T}_F$ tensor is so that it is symmetric (as the other two tensors) in the $i,j$ indexes and hence computing the Jacobian of $\textbf{C}$ will be simpler. On the other hand, in the presence of corners special care must be taken in order to avoid shearing at those points; that is why we include the corner's indices in the definition of $\textbf{T}_F$ (see Figure \ref{fig:forma_fundamental}). For a pseudo-algorithm for the computation of the $3$ tensors, see Algorithm \ref{algo_tensor}.

\subsection{Constraints evaluation}

This way, evaluating all the $n_c$ constraints simultaneously reduces to computing the dot product matrix $\textbf{P}(t)\in\mathbb{R}^{n\times n}$ defined by $P_{ij}(t) = \langle p_i(t),p_j(t)\rangle$ and then doing a tensor product (see Equation \ref{weight_galerkin}):

\begin{equation}\label{eq_cons}
\textbf{C}(\bm{\varphi}(t)) = \textbf{T}_{E,F,G}\otimes\textbf{P}(t) - \textbf{C}_0
\end{equation}
where we have grouped the previous tensors in just one $\textbf{T}_{E,F,G} = [\textbf{T}_E;\textbf{T}_F;\textbf{T}_G]\in\mathbb{R}^{n_c\times n\times n}$, and of course $\textbf{C}_0 = \textbf{T}_{E,F,G}\otimes\textbf{P}(0)$. With this definition, computing the Jacobian matrix $\nabla\textbf{C}\in\mathbb{R}^{n_c\times 3n}$ is very easy (because the tensors are symmetric in the last two indexes):
\begin{equation}\label{gradC}
\nabla\textbf{C}(\bm{\varphi}(t)) = 
2\cdot\left[\textbf{T}_{E,F,G}\otimes\textbf{x}(t),\textbf{T}_{E,F,G}\otimes\textbf{y}(t),\textbf{T}_{E,F,G}\otimes\textbf{z}(t)\right],
\end{equation}
recall that $\bm{\varphi}(t) = (\textbf{x}(t),\textbf{y}(t),\textbf{z}(t))^\intercal\in\mathbb{R}^{3n}$.

\subsection{Practical implementation}

We now list some considerations to take into account in a practical implementation of our method:
\smallskip

\noindent- As mentioned before, the constraints (\ref{modelo_grinspun}) are added to $\textbf{C}$ to impose that every edge of all boundary curves are preserved. Also, we remark again that the tensor constraints are only applied for interior nodes, except in the presence of corners, where we include their indices in the definition of $\textbf{T}_F$ in order to avoid shearing there.

\noindent- The tensor $\textbf{T}_{E,F,G}$ needs to be computed only once at the beginning of the simulation because it is time independent. All the integrals involved are evaluated exactly using standard \textit{Gaussian quadratures} (see \cite{Braess:2007:FE} for details) as shown in Algorithm \ref{algo_tensor}. 
\smallskip

\noindent- In order to be efficient we first calculate the Jacobian (according to Equation \ref{gradC}) and then put $\textbf{C}(\bm{\varphi}) = \frac{1}{2}\nabla\textbf{C}(\bm{\varphi})\cdot\bm{\varphi} - \textbf{C}_0$, so the matrix $\textbf{P}(t)$ is actually never computed.
\smallskip

\noindent- The tensor $\textbf{T}_{E,F,G}$ is highly sparse due to the fact that most products of the $3$ indicator functions (Equation \ref{weight_galerkin}) are zero. Nevertheless, it is not difficult to see that it has of the order of $\sim150n$ non-zero elements.

\section{Equations of motion}
Finally we can write down the Lagrangian (kinetic minus potential energies) of our discretized surface:

\begin{align*}
\mathcal{L}(\bm{\varphi}(t),\dot{\bm{\varphi}}(t))&=\frac{\rho}{2} \bm{\dot{\varphi}}(t)^\intercal\cdot \textbf{M} \cdot\bm{\dot{\varphi}}(t) - \rho \mathcal{G}^\intercal\cdot \textbf{M}\cdot \bm{\varphi}(t)\\
&-\frac{\kappa}{2} \bm{\varphi}(t)^\intercal\cdot \textbf{K} \cdot\bm{\varphi}(t)-\bm{\lambda}(t)^\intercal\cdot\textbf{C}(\bm{\varphi}(t)),
\end{align*}
where
\begin{enumerate}
	\item $\rho>0$ is the density of the cloth, $\textbf{M}$ is the augmented mass matrix and $\mathcal{G} = (0,\dots,0|0,\dots,0|g,\dots,g)^\intercal$ where $g = 9.8m/s^2$ is gravity,
	\item the stiffness matrix (we are using the isometric bending model described in \cite{Bergou:2006:QBM}) is $\textbf{K} = \textbf{L}^\intercal\textbf{M}\textbf{L}$
	where $\textbf{L}$ is an approximation of the point-wise Laplacian and $\kappa>0$ is a bending constant,
	\item and finally $\bm{\lambda}(t)$ are the Lagrange multipliers ensuring inextensibility (at the interior nodes and boundaries) and other possible positional constraints (e.g. the corners of the textiles to be manipulated) included.
\end{enumerate}

Thus, we get as Euler-Lagrange equations \cite{Taylor:2005:CM} the following ODE system:
\begin{equation}\label{ODE_sys}
\begin{cases}
\rho\textbf{M}\ddot{\bm{\varphi}} = \textbf{F}_{\rho} - \kappa\textbf{K}\bm{\varphi} - \textbf{D}\dot{\bm{\varphi}} - \nabla\textbf{C}(\bm{\varphi})^\intercal\bm{\lambda}\\
\textbf{C}(\bm{\varphi}) = 0
\end{cases}
\end{equation}
where $\textbf{F}_{\rho} = -\rho\textbf{M}\mathcal{G}$ is the force of gravity and we have added Rayleigh damping: $\textbf{D}= \alpha\textbf{M} + \beta\textbf{K},$ where $\alpha$ and $\beta$ are positive parameters \cite{Zienkiewicz:2005:FEM}. 

\smallskip

Notice how our inextensibility assumption reduces greatly the number of physical parameters of the model (we do not have shearing or stretching parameters and their respective dampings). Nevertheless, cloth dynamics can be very complicated and there is one important factor we are not accounting for in the previous equations: air resistance. Although there exist some simplified models \cite{Mao2009}, aerodynamic forces are difficult to model, especially near the boundaries of cloth, because turbulences appear and the nonrigid response of cloth to such turbulences is way more unpredictable than that of a rigid, even vibrating, body (see 
\cite{Shelley:2011}).   

\subsection{Equivalence principle and aerodynamics}


The equivalence principle states that for any object its inertial mass is equal to its gravitational mass, it means that an object in vacuum of inertial mass $m$ would fall freely under the action of a force of magnitude $mg$, where $g$ is gravity. Nevertheless, experiments show that in the presence of air the free falling velocities depend on the (shape of the) object at hand. In our experiments we have found that aerodynamic effects can be modeled without explicitly including them by allowing inertial and gravitational masses to be different. Although this is not true in the physical world, the consideration of the
two masses as virtual in our model allows us to account for all aerodynamic contributions to cloth dynamics (drag, lift, turbulences at the boundaries) with great accuracy. Hence, we put $F_{\delta} = \delta\textbf{M}\mathcal{G}$, setting $\delta$ as a new parameter to be fitted. In Table \ref{tabla:0} we summarize the meaning of all the parameters of our model.

\begin{table}[h!]
	\centering
	\begin{tabular}{|c|l|}
		\hline
		\multicolumn{1}{|l|}{Parameter} & \multicolumn{1}{c|}{Meaning}                     \\ \hline
		$\rho$                          & Density (inertial mass)      \\
		$\delta$                        & Virtual (gravitational)  mass           \\
		$\kappa$                        & Bending/stiffness            \\
		$\alpha$                        & Damping of slow oscillations \\
		$\beta$                         & Damping of fast oscillations \\
		\hline
	\end{tabular}
	\caption{Physical parameters of the model and their meaning.}
	\label{tabla:0}
\end{table}

\section{Numerical integration and contact}
As usual, we approximate $\bm{\varphi}(t)$ and $\dot{\bm{\varphi}}(t)$ with $\{\bm{\varphi}^0,\bm{\varphi}^1,\dots\}$ and $\{\dot{\bm{\varphi}}^0,\dot{\bm{\varphi}}^1,\dots\}$, where $\bm{\varphi}^n$ and $\dot{\bm{\varphi}}^n$ are the position and velocities of the nodes of the mesh at time $t_n = n\cdot dt$ and $dt > 0$ is the size of the time step. Using an implicit Euler scheme to integrate equations (\ref{ODE_sys}) we obtain:
\begin{equation}\label{proj_int}
\begin{cases}
\bm{\varphi}^{n+1} = \bm{\varphi}_0^{n+1} -\textbf{M}^{-1}\nabla\textbf{C}(\bm{\varphi}^{n+1})^\intercal\bm{\lambda}^{n+1}\\
\textbf{C}(\bm{\varphi}^{n+1}) = 0,
\end{cases}
\end{equation}
where $\bm{\varphi}_0^{n+1}$ is the unconstrained step (which depends on $\bm{\varphi}^{n}$ and $\dot{\bm{\varphi}}^n$). Note that at time $t_n = n\cdot dt$ the only unknown in previous equations is $\bm{\varphi}^{n+1}$ (and $\bm{\lambda}^{n+1}$). 

\subsection{Fast projection algorithm}

Finding $\bm{\varphi}^{n+1}$ can be done solving the following  constrained optimization problem:
\begin{equation}\label{quad_problem}
\begin{cases}
\min_{\bm{\varphi}^{n+1}}\quad(\bm{\varphi}^{n+1}-\bm{\varphi}_0^{n+1})^\intercal\cdot\textbf{M}\cdot(\bm{\varphi}^{n+1}-\bm{\varphi}_0^{n+1}) \\
\text{s.t. }\quad\textbf{C}(\bm{\varphi}^{n+1}) = 0.
\end{cases}
\end{equation}
because equations (\ref{proj_int}) are the critical (stationary) points of the optimization problem. This is a quadratic program with quadratic constraints. In order to make it computationally tractable, it is approximated by a \textit{sequence} of quadratic programs with linear constraints: write $\bm{\varphi}_{j+1} = \bm{\varphi}_j + \Delta\bm{\varphi}_{j+1}$ and make the approximation $\textbf{C}(\bm{\varphi}_{j+1})= \textbf{C}(\bm{\varphi}_j + \Delta\bm{\varphi}_{j+1})\simeq \textbf{C}(\bm{\varphi}_{j}) + \nabla\textbf{C}(\bm{\varphi}_{j})\Delta\bm{\varphi}_{j+1}$, then the sequence is:
\begin{equation*}
\begin{cases}
\min_{\Delta\bm{\varphi}_{j+1}}\quad\Delta\bm{\varphi}_{j+1}^\intercal\cdot\textbf{M}\cdot\Delta\bm{\varphi}_{j+1} \\

\text{s.t. }\textbf{C}(\bm{\varphi}_{j}) + \nabla\textbf{C}(\bm{\varphi}_{j})\cdot\Delta\bm{\varphi}_{j+1} = 0,
\end{cases}
\end{equation*}
being the initial point $\bm{\varphi}_0 = \bm{\varphi}_0^{n+1}$. This is called the \textit{fast projection algorithm} \cite{Goldenthal:2007:ESI}. Each of these quadratic programs can be reduced to solving a linear system and we iterate until the constraints are satisfied to a given relative tolerance (usually $0.1\%$). In the case we want to simulate really fine meshes (of the order of thousands of nodes), more efficient algorithms (e.g. \cite{Tour2015}) can be used.

\subsection{Contact and collisions with an obstacle}

For its application in the real world, we need to include into our model collisions of the cloth with a fixed object (e.g. a table). We will assume this obstacle is given by an implicit equation $\textbf{H}(\bm{\varphi}) = 0$ with a well defined outwards normal $\nabla\textbf{H}(\bm{\varphi})$ (almost everywhere). We can model collisions with the aim of non-smooth dynamics \cite{Kun00,Zho93}, and include these new forces in every iteration of the \textit{fast projection algorithm} described earlier. Signorini's contact model reads:
\begin{equation}\label{eq_colisiones}
\begin{cases}
\textbf{M}\ddot{\bm{\varphi}} = \textbf{F}(\bm{\varphi},\dot{\bm{\varphi}}) - \nabla\textbf{C}(\bm{\varphi})^\intercal\bm{\lambda} + \nabla\textbf{H}(\bm{\varphi})^\intercal\bm{\gamma},\\
\textbf{C}(\bm{\varphi}) = 0,\\
\textbf{H}(\bm{\varphi}) \geq 0,\quad \bm{\gamma}\geq 0,\quad \bm{\gamma}^\intercal\cdot \textbf{H}(\varphi) = 0,
\end{cases}
\end{equation}
where we have grouped in $ \textbf{F}(\bm{\varphi},\dot{\bm{\varphi}})$ damping, gravity, etc. Now the system is non-smooth, that is why we need to use a first-order scheme \cite{Kun00}. To integrate it numerically, we put again  $\bm{\varphi}_{j+1} = \bm{\varphi}_j + \Delta\bm{\varphi}_{j+1}$ where the initial point is the unconstrained step $\bm{\varphi}_0 = \bm{\varphi}_0^{n+1}$ given by an implicit Euler scheme. Also, we write:
\begin{equation*}
\textbf{H}(\bm{\varphi}_{j+1})= \textbf{H}(\bm{\varphi}_j + \Delta\bm{\varphi}_{j+1})\simeq \textbf{H}(\bm{\varphi}_{j}) + \nabla\textbf{H}(\bm{\varphi}_{j})\Delta\bm{\varphi}_{j+1},
\end{equation*}
and then solve iteratively the following sequence of quadratic programs with linear equality and inequality constraints:
\begin{equation*}
\begin{cases}
\min_{\Delta\bm{\varphi}_{j+1}}\tfrac{1}{2}\Delta\bm{\varphi}_{j+1}^\intercal\cdot\textbf{M}\cdot\Delta\bm{\varphi}_{j+1}\\
\textbf{C}(\bm{\varphi}_{j}) + \nabla\textbf{C}(\bm{\varphi}_{j})\Delta\bm{\varphi}_{j+1} = 0,\\
\textbf{H}(\bm{\varphi}_{j}) + \nabla\textbf{H}(\bm{\varphi}_{j})\Delta\bm{\varphi}_{j+1} \geq 0.
\end{cases}
\end{equation*}

Note that the critical points (compare to (\ref{eq_colisiones})) of the previous quadratic problems are:

\begin{equation*}
\begin{cases}
\textbf{M}\cdot\Delta\bm{\bm{\varphi}}_{j+1} = -\nabla\textbf{C}(\bm{\varphi}_{j})^\intercal\Delta\bm{\lambda}_{j+1} + \nabla\textbf{H}(\bm{\varphi}_{j})^\intercal\Delta\bm{\gamma}_{j+1},\\
\textbf{C}(\bm{\varphi}_{j}) + \nabla\textbf{C}(\bm{\varphi}_{j})\Delta\bm{\varphi}_{j+1} = 0,\\
\textbf{H}(\bm{\varphi}_{j}) + \nabla\textbf{H}(\bm{\varphi}_{j})\Delta\bm{\varphi}_{j+1} \geq 0,\\ 
\Delta\bm{\gamma}_{j+1}\geq 0,\quad \Delta\bm{\gamma}_{j+1}^\intercal\cdot \left[\textbf{H}(\bm{\varphi}_{j}) + \nabla\textbf{H}(\bm{\varphi}_{j})\Delta\bm{\varphi}_{j+1}\right] = 0.
\end{cases}
\end{equation*}

\section{Evaluation and results}

\subsection{Locking test}
\begin{figure}[]
	\centering
	\includegraphics[width=0.5\linewidth]{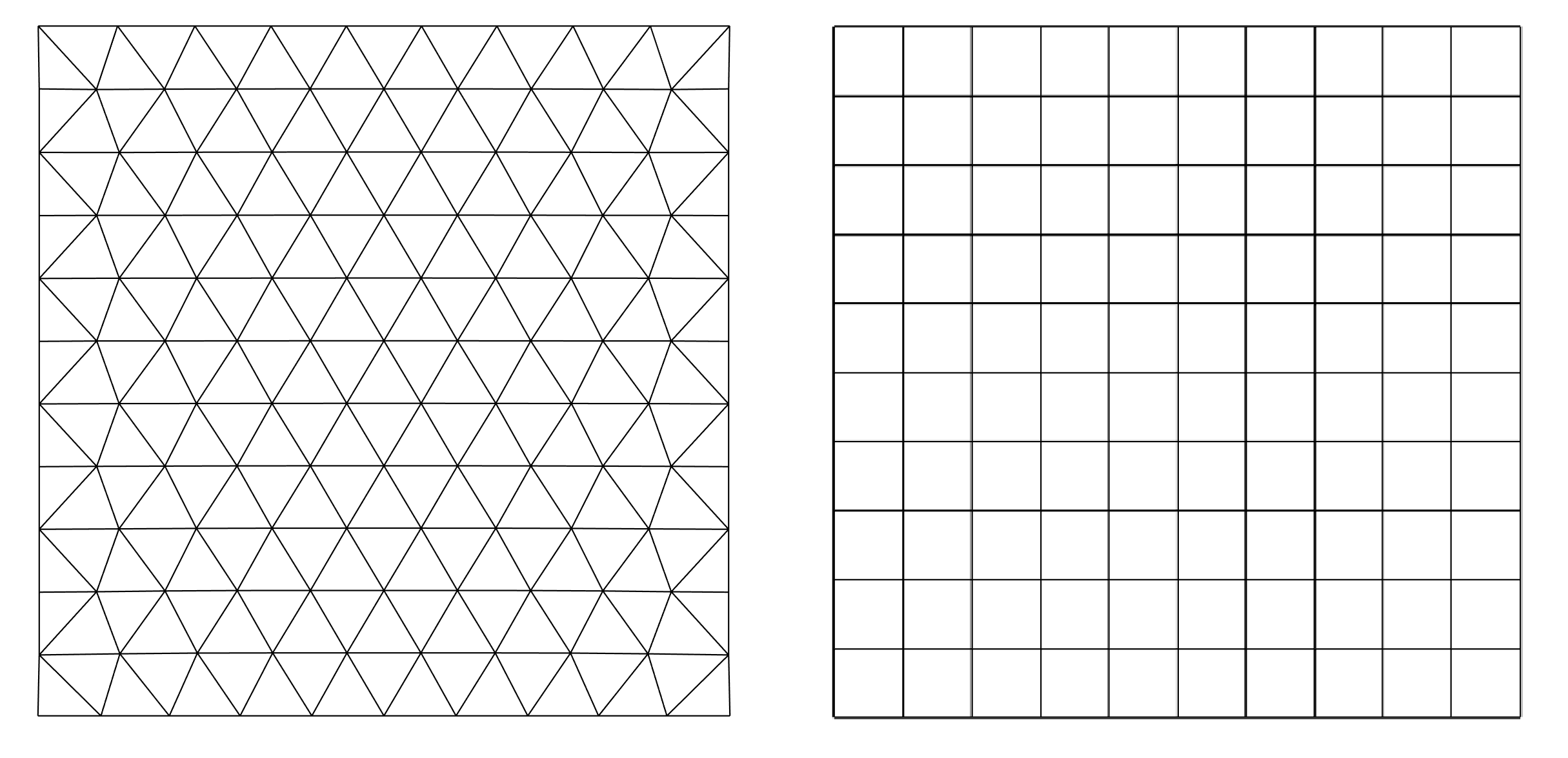}
	\caption{\label{triang_quad} Different meshes used for the locking test. The triangular meshes (left) are irregular whereas the quadrilateral ones (right) are uniform.}
\end{figure}

With this experiment we intend to show the locking-free nature of our model, and its stability with respect to mesh topology. We fix three corners of a flat sheet of cloth (with added random noise of standard deviation 3mm) of 1m by 1m, and let the fourth fall freely. We use four different meshings of the cloth: two with triangles and two with quadrilaterals (see Figure \ref{triang_quad}), but we keep the physical parameters fixed.


\begin{figure}[]
	\centering
	\includegraphics[width=1\linewidth]{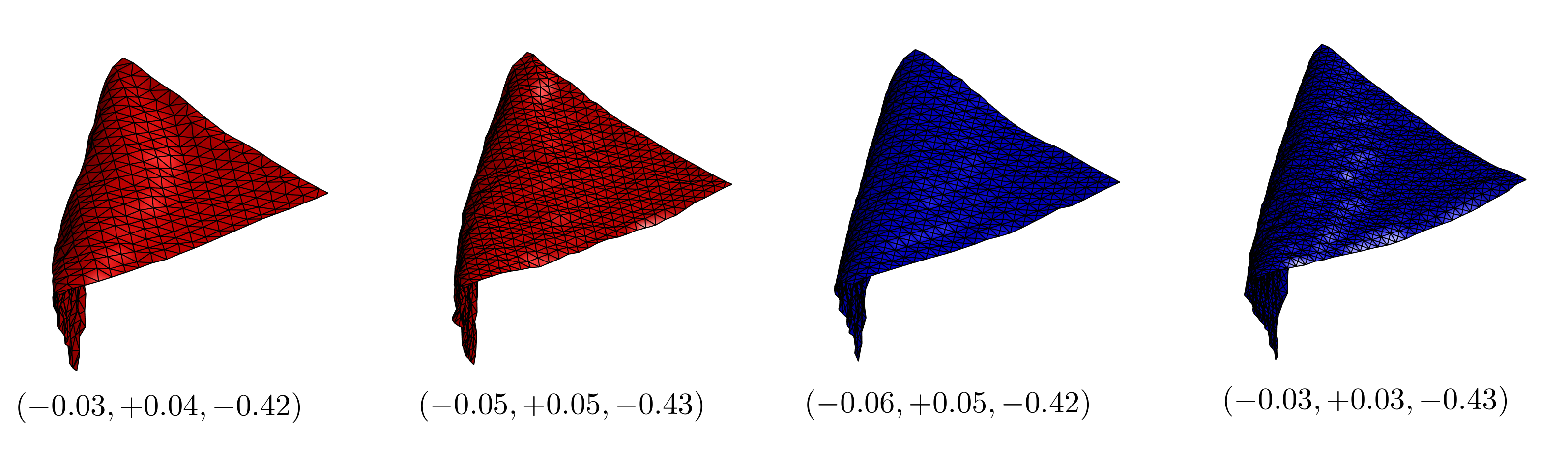}
	\caption{\label{four_mantas} Four different meshes used for the experiment: the red ones are made of triangles and the blue ones of quadrilaterals. From left to right the number of nodes are: 472, 941, 529, 900. On the bottom of each blanket, we display the euclidean position of the free corner.}
\end{figure}

The visual results of the experiment can be seen in Figure \ref{four_mantas}. The four blankets fold diagonally without any locking artifact. 

Moreover, in Table \ref{table:2} we compute the euclidean distance of the bottom corner of every textile with respect to each other. Note how, doubling the number of nodes, or changing from irregular triangles to regular quadrilaterals, does not alter the result within a margin of error of a few centimeters (recall that the sheets are 1m$\times$1m).

\begin{table}[]
	\centering
	\begin{tabular}{c|c|c|c|c|}
		\cline{2-5}
		                        & \multicolumn{1}{l|}{472 nodes $\triangle$} & \multicolumn{1}{l|}{941 nodes $\triangle$} & \multicolumn{1}{l|}{529 nodes  $\square$} & \multicolumn{1}{l|}{900 nodes  $\square$} \\ \hline\hline
		\multicolumn{1}{|l||}{472 nodes $\triangle$} & 0                                        & 2.45 cm                                  & 3.16 cm                                & 1.41 cm                                \\ \hline
		\multicolumn{1}{|l||}{941 nodes $\triangle$} & -                                  & 0                                        & 1.41 cm                                & 2.83 cm                                \\ \hline
		\multicolumn{1}{|l||}{529 nodes  $\square$}   & -                                  & -                                  & 0                                      & 3.74 cm                                \\ \hline
		\multicolumn{1}{|l||}{900 nodes  $\square$}   & -                                  & -                                  & -                                & 0                                      \\ \hline
	\end{tabular}\caption{Distances (cm) between the bottom corner of the cloths (of dimensions 1m$\times$1m).}
\label{table:2}
\end{table}

\subsection{Cusick's test}\label{section_cusick}

\begin{figure}[H]
	\centering
	\includegraphics[width=0.35\linewidth]{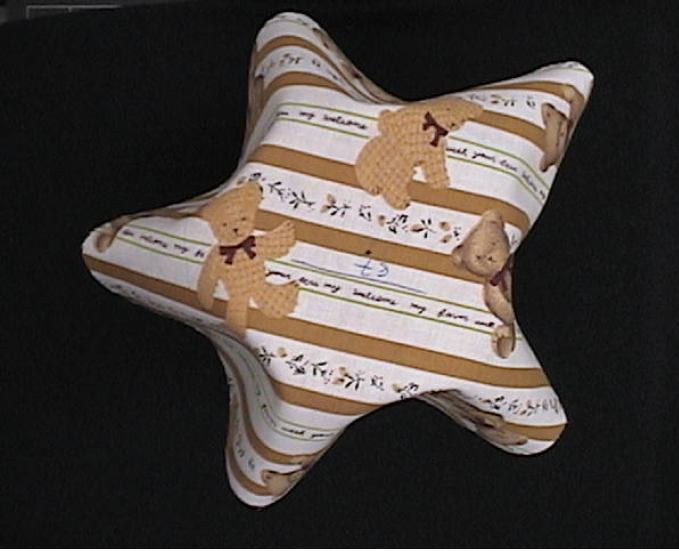}
	\caption{\label{real_cusick} Photo of a circular cloth draping on top of a circular table. Image taken from \cite{Gider2004}.}
\end{figure}

In this section we explain how to simulate Cusick's test \cite{Hu04} using our simulator. This experiment consists in letting a circular cloth of radius 15cm drape on top of an also circular table of radius 9cm (see Figure \ref{real_cusick}) with their centers aligned. 

\begin{figure}[htb]
	\centering
	\includegraphics[width=0.65\linewidth]{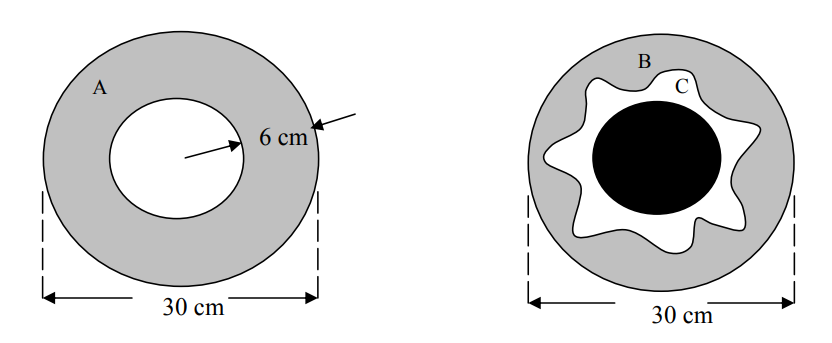}
	\caption{\label{esquema_cusick} After the cloth has draped on top of the table, the drape coefficient is calculated as the ratio between the area of the plane projection of the cloth $C$, divided by the area of the flat ring $A$ of width 6cm. Image taken from \cite{Gider2004}.}
\end{figure}

Then, one computes the area of the plane projection of the draped cloth $C$ and divides it by the area of the flat ring $A$ of width 6cm (see Figure \ref{esquema_cusick}). 

This is called the \textbf{drape coefficient} of the cloth:

\begin{equation}
\text{DC}\; \% = 100\times\frac{C}{A}.
\end{equation}

It is known that this coefficient depends heavily on the stiffness of the cloth  \cite{Hu04}. This can be understood intuitively: a completely rigid sheet of metal would not bend and would have DC equal to $100\%$, whereas a really light material (e.g. silk) has a DC pretty close to 0. The measurement of this coefficient is not trivial: it usually requires a dedicated machine (Cusick's Drape Tester \cite{K82}), and variation of the measured coefficient for the same cloth is common, so several specimens of the same textile have to be employed and then their DC's averaged \cite{Gider2004}. 

\begin{figure}[htb]
	\centering
	\includegraphics[width=0.9\linewidth]{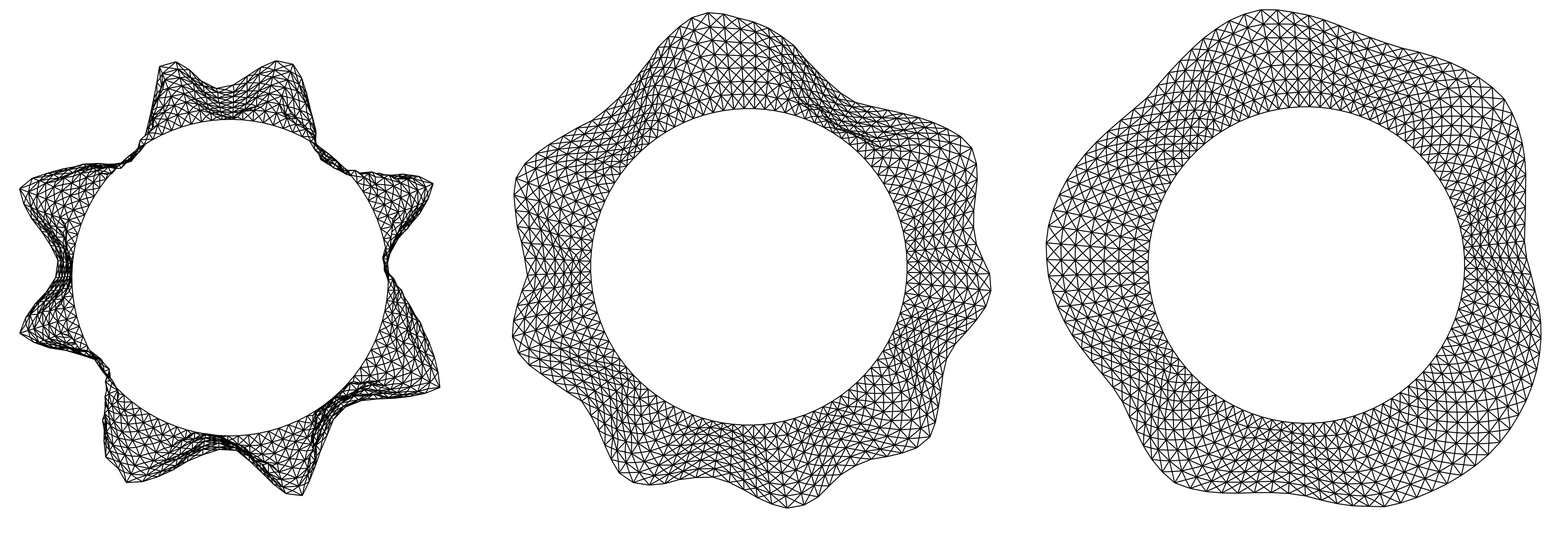}
	\caption{\label{inextCusick}Simulation of Cusick's test with the inextensible model and a mesh of 768 nodes. From left to right we vary the stiffness of the cloth and get respectively DCs of $26.1\%, 54.9\%, 77.2\%$. They correspond to \textit{peach-skin} polyester (low stiffness), \textit{imitation} wool (medium) and \textit{tussore} cotton (high).}
\end{figure}

\begin{figure}[H]
	\centering
	\includegraphics[width=0.7\linewidth]{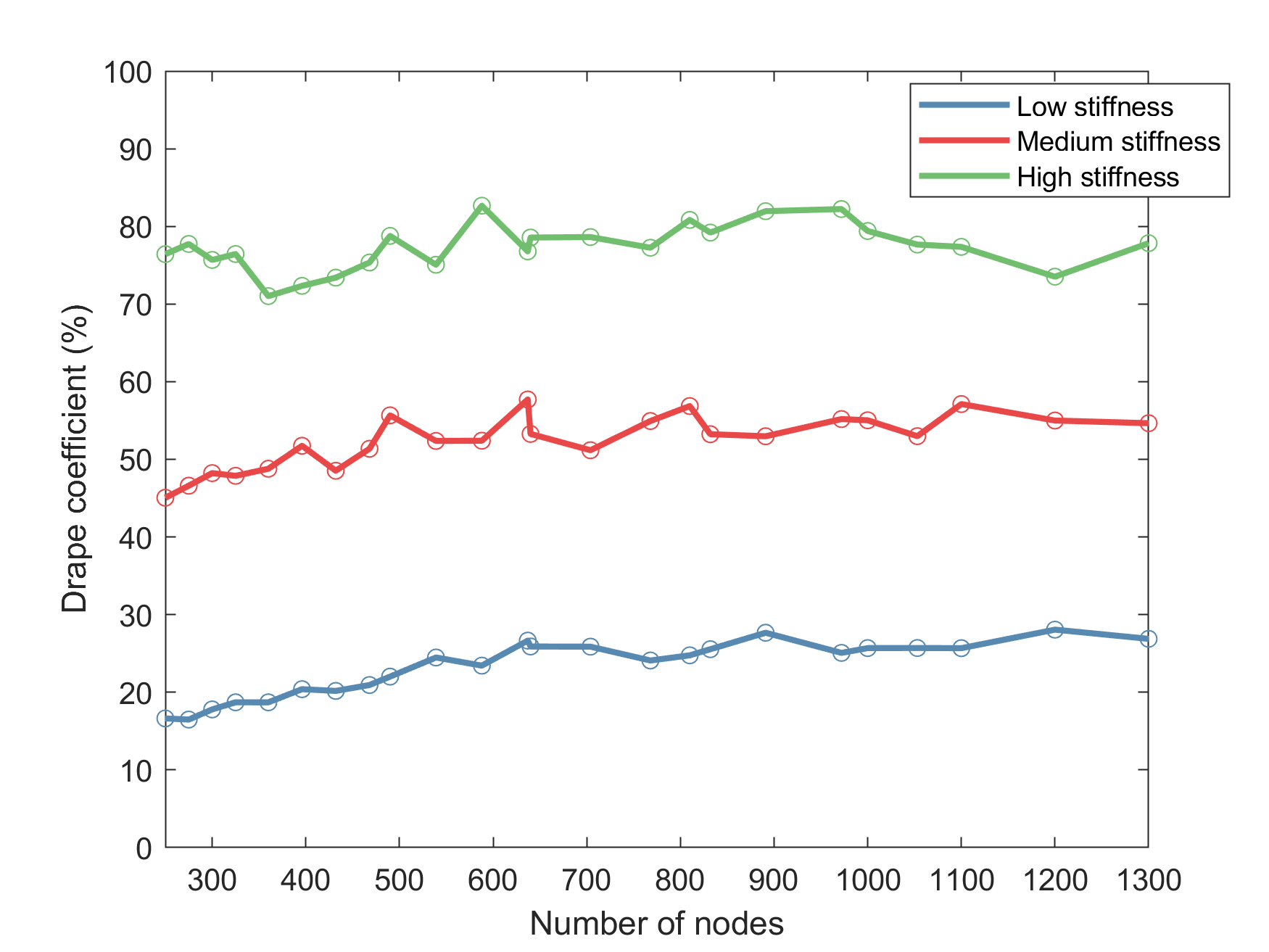}
	\caption{\label{res_cusick} Computation of the drape coefficient for 3 stiffness values (low, medium and high) and $24$ different meshes of cloth from $250$ to $1300$ nodes. The three mean values for the computed DCs are $23.2\%,\;52.4\%,\;77.3\%$.}
\end{figure}

Naturally, this has caused the creation of alternative simulation methods (e.g. \cite{Gan95}). In order to simulate Cusick's test with our model, we  quadrangulate the ring $A$ of Figure \ref{esquema_cusick}, we fix in space its inner nodes and let gravity act (see Figure \ref{inextCusick}). In order to study the dependency of the DC with respect to mesh resolution, in Figure \ref{res_cusick} we plot the DC for three stiffness values (low, medium and high) and $24$ different meshes of the ring $A$ ranging from $250$ nodes to $1300$. The three mean values for the computed DCs are $23.2\%,\;52.4\%,\;77.3\%$ and they approximately correspond to \textit{peach-skin} polyester (low stiffness), \textit{imitation} wool (medium stiffness) and \textit{tussore} cotton (high stiffness) (see the Appendix of \cite{Gider2004}: IDs: $38,37,22$). The figure shows that the computation of the DC with our model is very stable (especially from 700 nodes on), and hence the model has a robust behavior with respect to mesh resolution. As we already said, this is very relevant for robotic applications, where the use of coarse meshes becomes a necessity because of performance constraints. 

\subsection{Tank-top simulation}
\begin{figure}[H]
	\centering
	\includegraphics[width=1\linewidth]{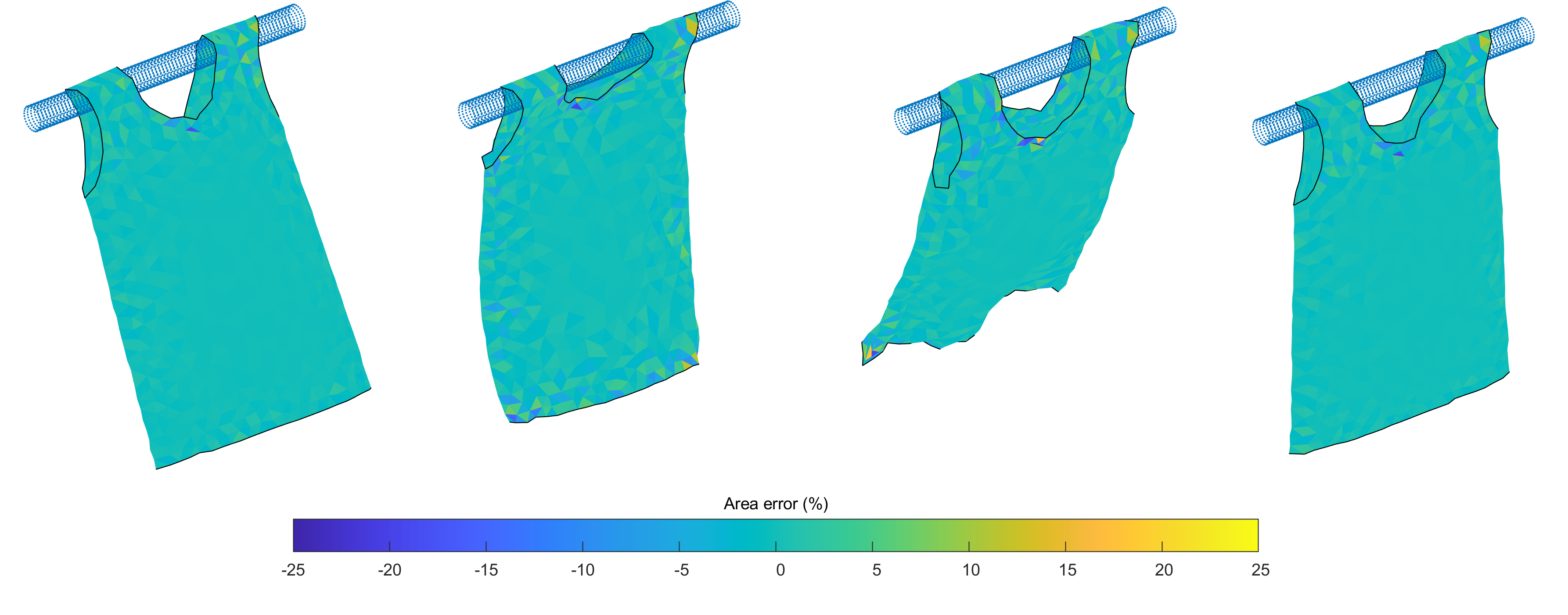}
	\caption{\label{camiseta_mov} Simulation with the inextensible model of the shaking of a triangulated tank-top. At each time instant ($t= 1.1s,\; 1.75s,\; 2.5s,\; 3.5s$) we plot the area error with sign (\ref{error_signo}) of each individual triangle.}
\end{figure}

With this experiment we aim to show the ability of our model to simulate complex topologies (e.g. a tank-top shirt, see Figure \ref{camiseta_mov}) and study different area errors. We track the variation of area of the garments through time. We think area is a good measure since it is a physical property of cloth and not a mesh-dependent metric (e.g. stretching of the edges of the quadrilaterals). We compute the area of each element of the discretized cloth using its local parametrization (\ref{quad_param}). 

We use three area errors: 
\begin{enumerate}
	\item \textit{Total area error}:
	\begin{equation}\label{error_area}
	e_t(t) = \frac{|A(t)-A_0|}{A_0},
	\end{equation}
	where $A_0$ is the total area of the garment at time $t=0$ and $A(t)$ is its area at time $t>0$.
    \item \textit{Mean element error}:
	\begin{equation}\label{error_medio}
	e_m(t) = \frac{1}{n_e}\sum_{e}\frac{|a_e(t)-a_e(0)|}{a_e(0)},
	\end{equation}
	where $a_e(t)$ is the area of element $\Omega_{e}$ at time $t\geq0$.
	\item \textit{Dispersion}:
	\begin{equation}\label{dispersion_error}
	d_a(t) = e_m(t) + 2\sqrt{\Var_e\left(\frac{|a_e(t)-a_e(0)|}{a_e(0)}\right)}.
	\end{equation}
\end{enumerate}

We simulate the shaking with a \textit{hanger} of a meshed tank-top with $1676$ triangles (see Figure \ref{camiseta_mov}) during 4.5s. There, we also plot the area error with sign of each individual triangle for each time instant:
\begin{equation}\label{error_signo}
e(t) = \frac{a_e(t)-a_e(0)}{a_e(0)}.
\end{equation}

\begin{figure}[htb]
	\centering
	\includegraphics[width=0.8\linewidth]{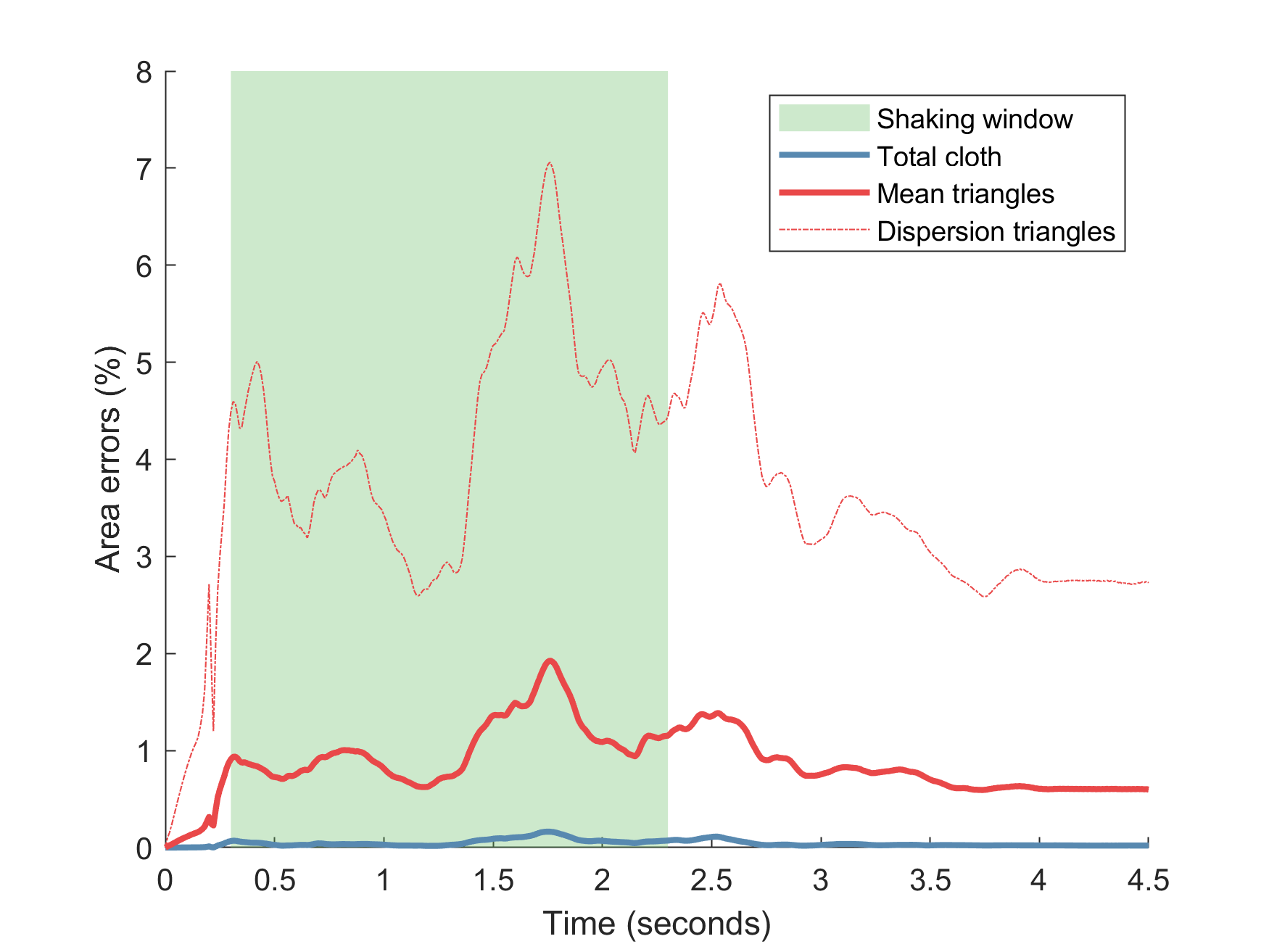}
	\caption{\label{errores_area} Plot of the time-dependent errors: total area (\ref{error_area}), mean element (\ref{error_medio}) and  dispersion (\ref{dispersion_error}). The total area error is almost zero. The mean area error is higher, but still very low, around $1\%$.}
\end{figure}

It is interesting to notice the concentration of error near the boundaries of the cloth and at the points where the tank-top makes contact with the moving cylinder. Nevertheless, note that when a triangle is stretched (color yellow), next to it appears a triangle that contracts (color blue). This makes that the total area error be very small (see Figure \ref{errores_area}). Finally in Figure \ref{errores_area}, we plot the three area errors (\ref{error_area}),(\ref{error_medio}) and (\ref{dispersion_error}). The total area error is very low during all the simulation, almost zero. The mean area error is higher, but still very low (it peaks around $t=1.75s$, the second frame in Figure \ref{camiseta_mov}), keeping itself around $1\%$. The dispersion error on the other hand, tells us that if the distribution of the errors was normal, the error of $95\%$ all the triangles of the mesh would be under $7\%$ during all the simulation. 

\section{Experimental validation} \label{s:expvalidation}
\subsection{Camera data}\label{camera}
\begin{figure}[htb]
	\centering
	\includegraphics[width=0.72\linewidth]{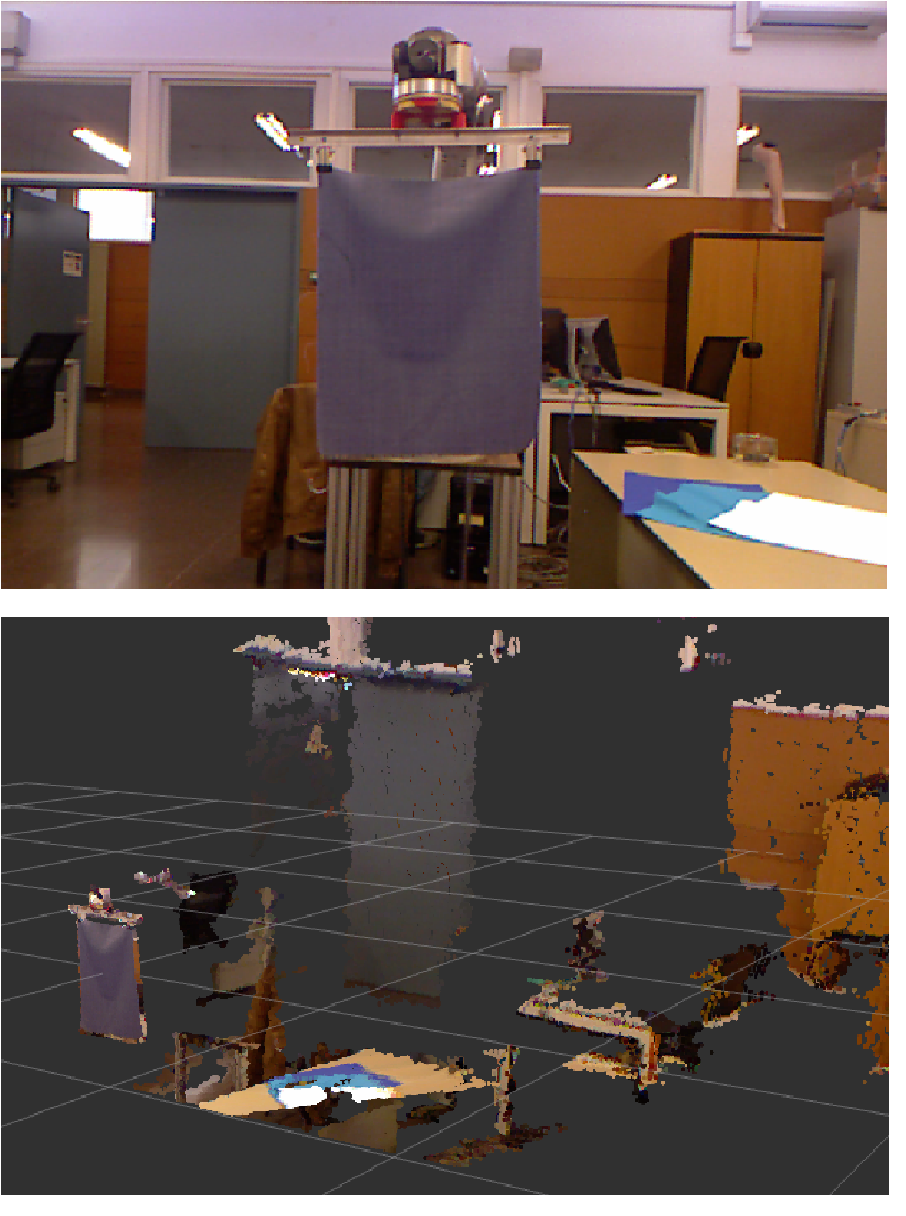}
	\caption{\label{camara}Point-cloud obtained using a depth camera (bottom) and flat image (top). Note that the camera only gets the depth of the objects visible to it, all the rest (e.g. the robot behind the cloth) is occluded.}
\end{figure}

Comparison with reality is performed by recording in the laboratory the motion of a piece of cloth when subjected to exactly the same movements of the robotic arm controlling it as specified in the simulations with our model. The real motion is captured by a depth camera Kinect XB360-607 (see Figure \ref{camara}). The garment is a rectangular piece of cloth, with its two upper corners fixed to a rigid hanger, which is affixed itself to a robotic arm Barrett WAM. To keep the garment completely in view of the camera, the robot moves it along the depth axis of the camera (axis $x$ in our reference) following a curve of equation
\begin{equation}\label{osc_motion}
(x(t),y(t),z(t)) = (A\cos(2\pi f t) + c,y_0,z_0).
\end{equation}
The two corners of the garment fixed to the robot arm follow the same oscillation, with different but constant values $y_1,y_2$ instead of $y_0$.

\begin{figure}[htb]
	\centering
	\includegraphics[width=0.4\linewidth]{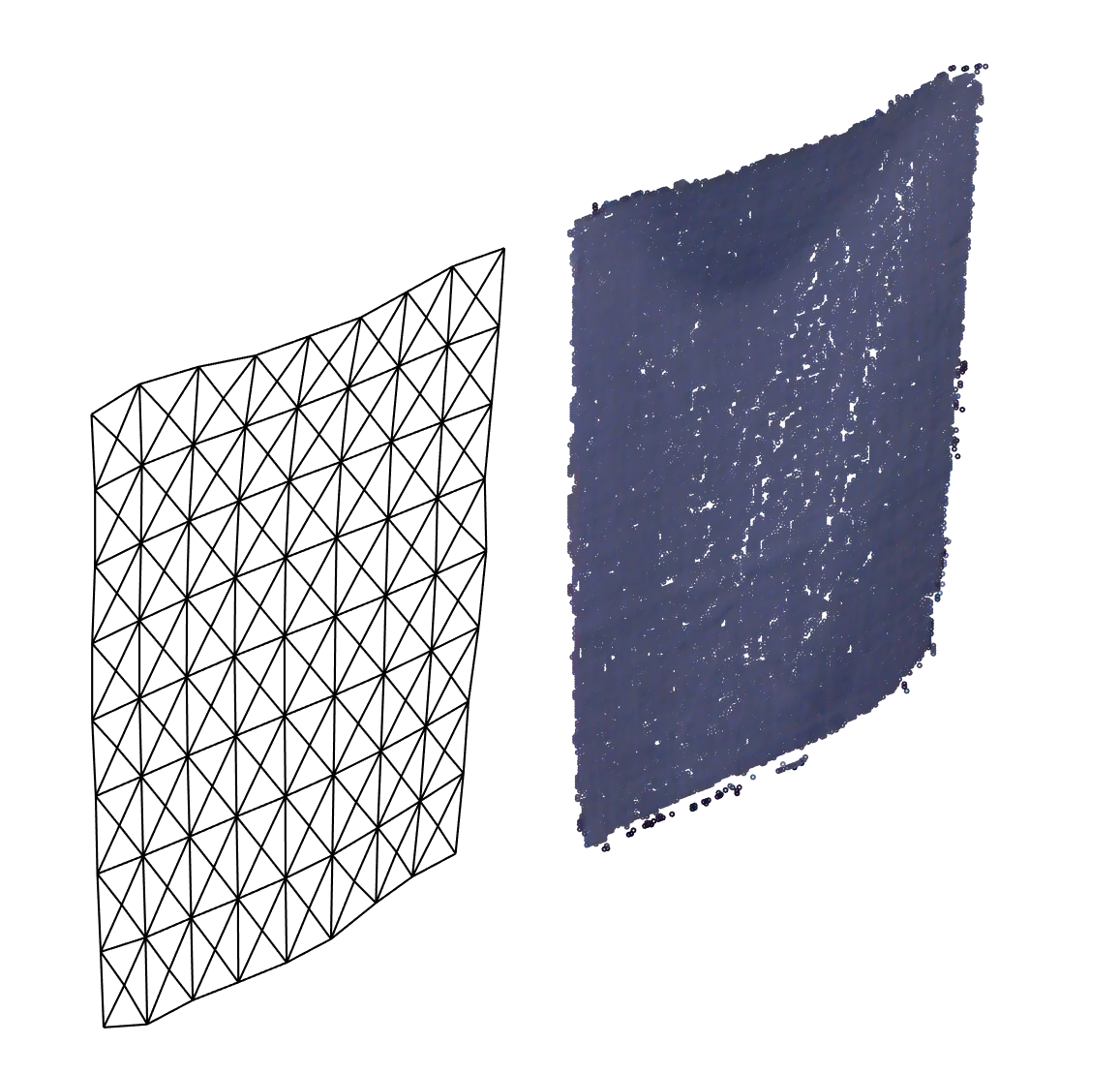}
	\caption{\label{nube} Quadrilateral meshing (left) of one of the frames of the filtered and de-noised point-cloud (right). Each quadrilateral is divided into triangles for plotting purposes.}
\end{figure}

Two motions are recorded: the \textit{slow} one has $A = 0.15$m and $f = 0.3$Hz, while the \textit{fast} one has $A = 0.075$m and $f = 0.6$Hz. Surrounding objects are filtered using planes for each recorded frame, and outlying points are detected on the basis of distance and removed. Only garments with a homogeneous color are tested, so we can remove further noise through use of a color filter. After the point cloud has been thus filtered, it is meshed using the algorithm described in \cite{Hormann:2000:QR} (see Figure \ref{nube}). The model that we are validating is continuous, producing simulations which are very stable under remeshing. Because of this, it suffices to select as a baseline for all the experiments a coarse $9\times9$ mesh. 

The robotic arm follows the oscillation of Eq. \ref{osc_motion} for $10$ seconds. The camera starts recording $1$ second before the start, and finishes $3$ seconds after the end of the robotic motion, for $14$ seconds of recorded garment motion. The original recording has its frames at irregular time steps. We replace them through linear interpolation to obtain the position of the point cloud each $dt = 0.01$ seconds.

\subsection{Parameter fitting}	
We will be adjusting only 2 parameters: the damping parameter $\alpha$ and the (virtual) gravitational mass $\delta$. The other parameters in Table \ref{tabla:0} are set to $0$ except for $\rho = 1$. In Section \ref{camera} we explained how to obtain a sequence of positions of the nodes of the real cloth $\{\bm{\phi}^0,\bm{\phi}^1,\dots,\bm{\phi}^m\}$. If we integrate numerically Eq. ($\ref{ODE_sys})$ using the same trajectories (\ref{osc_motion}) for the two upper corners, we get a sequence $\{\bm{\varphi}^0(\delta,\alpha),\bm{\varphi}^1(\delta,\alpha),\dots,\bm{\varphi}^m(\delta,\alpha)\}$ (where $\bm{\varphi}^0 = \bm{\phi}^0$) for each value of the two parameters. Hence, a natural metric is:
\begin{equation}\label{min_cuad}
L(\delta,\alpha) = \sum_{i=1}^{\left[\tfrac{m}{2}\right]}||\bm{\varphi}^i(\delta,\alpha) - \bm{\phi}^i||^2,
\end{equation}
where $||\cdot||$ is the $L^2$ norm with respect to the matrix $\textbf{M}$, and we only use the first half (7 seconds) of the recorded frames to perform the fitting. Finally, we minimize $L$ using a derivative-free algorithm (the Nelder-Mead Simplex Method) and find the optimal values of $\delta,\alpha$. The resulting parameters are shown in Table \ref{table:1}. We will analyze the evolution of the time-dependent absolute error:


\begin{equation}\label{error_rel}
e_i(\delta,\alpha) = \sqrt{||\bm{\varphi}^i(\delta,\alpha) - \bm{\phi}^i||^2}.
\end{equation}
\begin{table}[h!]
	\centering
	\begin{tabular}{||c c c c c c c||} 
		\hline
		Material & $\delta_{slow}$ & $\delta_{fast}$ & $\alpha_{slow}$  & $\alpha_{fast}$ & $\bar{e}_{slow}$ & $\bar{e}_{fast}$\\ [0.5ex] 
		\hline\hline
		Paper & 0.32 & 0.37 & 1.40 & 2.50 & 0.45 cm & 0.41 cm\\ 
		\hline
		Cotton & 0.52 & 0.52 & 1.29 & 2.69 & 0.41 cm & 0.31 cm\\
		\hline
		Felt & 0.46 & 0.61 & 1.05 & 2.65 & 0.39 cm & 0.30 cm \\
		\hline
		Wool & 0.47 & 0.50 & 1.19 & 2.52 & 0.37 cm & 0.34 cm \\ [1ex] 
		\hline
	\end{tabular}
	\caption{Estimated parameters and mean absolute errors (cm).}
	\label{table:1}
\end{table}

In the last two columns of Table \ref{table:1} we show the average values of $e_i$ over the testing window (i.e. the last 7 seconds of the motion). Moreover, we use two times the standard deviation of the error at each node as a dispersion measure:

\begin{equation}\label{dispersion}
d_i(\delta,\alpha) = e_i(\delta,\alpha) + 2\sqrt{\Var_{\mathbb{R}}\left(|\bm{\varphi}^i(\delta,\alpha) - \bm{\phi}^i|_{\mathbb{R}^3}\right)}.
\end{equation}

\begin{figure}[!h]
	\centering
	\includegraphics[width=0.9\linewidth]{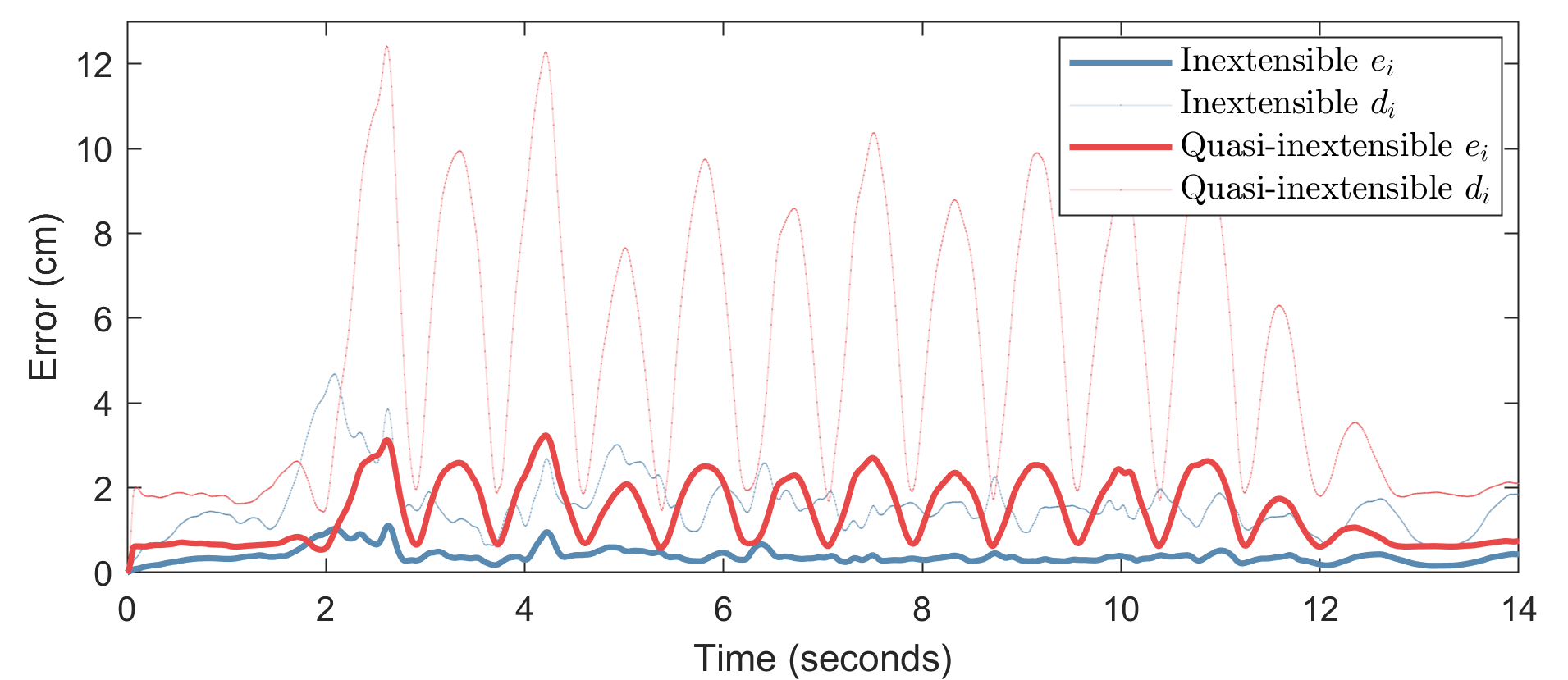}
	\caption{\label{comp_grispun}Comparison of the fast motion of light cotton between the inextensible and quasi-inextensible model's errors, with respect to the recorded motion. The mean absolute error for the inextensible model is 0.31cm, whereas for the quasi-inextensible is 1.40cm.}
\end{figure}

To have a baseline to compare with, we implement the quasi-inextensible model \cite{Goldenthal:2007:ESI}: in this formulation stretching is not allowed because all the edges of the quadrangulated cloth are constrained to maintain their length, and only shearing is permitted by using non-stiff diagonal springs. Therefore, we only fit Rayleigh's damping $\alpha$ and the stiffness of the shearing springs $\kappa$ for the $9\times 9$ mesh. The results for the fast motion of cotton are shown in Figure \ref{comp_grispun}. The average value of the absolute error (\ref{error_rel}) is $1.40$cm.

\subsection{Discussion of the results}
\begin{figure}[h!]
	\centering
	\includegraphics[width=0.85\linewidth]{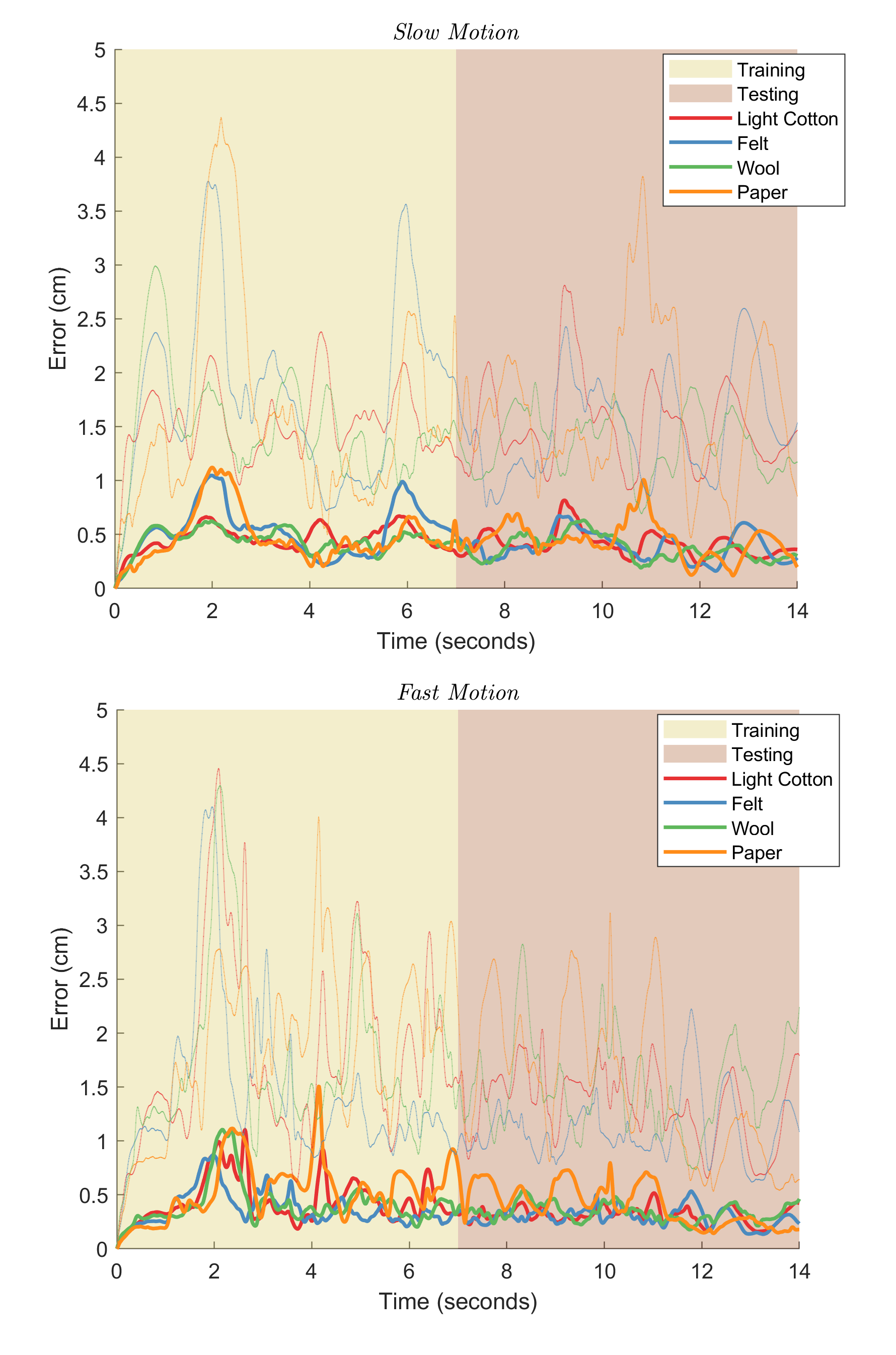}
	\caption{\label{movimientos} Absolute error (bottom curves) and dispersion (top curves) of the position of the simulated textile vs. the real motion for the slow movement (top) and the fast one (bottom).}
\end{figure}

\begin{figure}[H]
	\centering
	\includegraphics[width=0.9\linewidth]{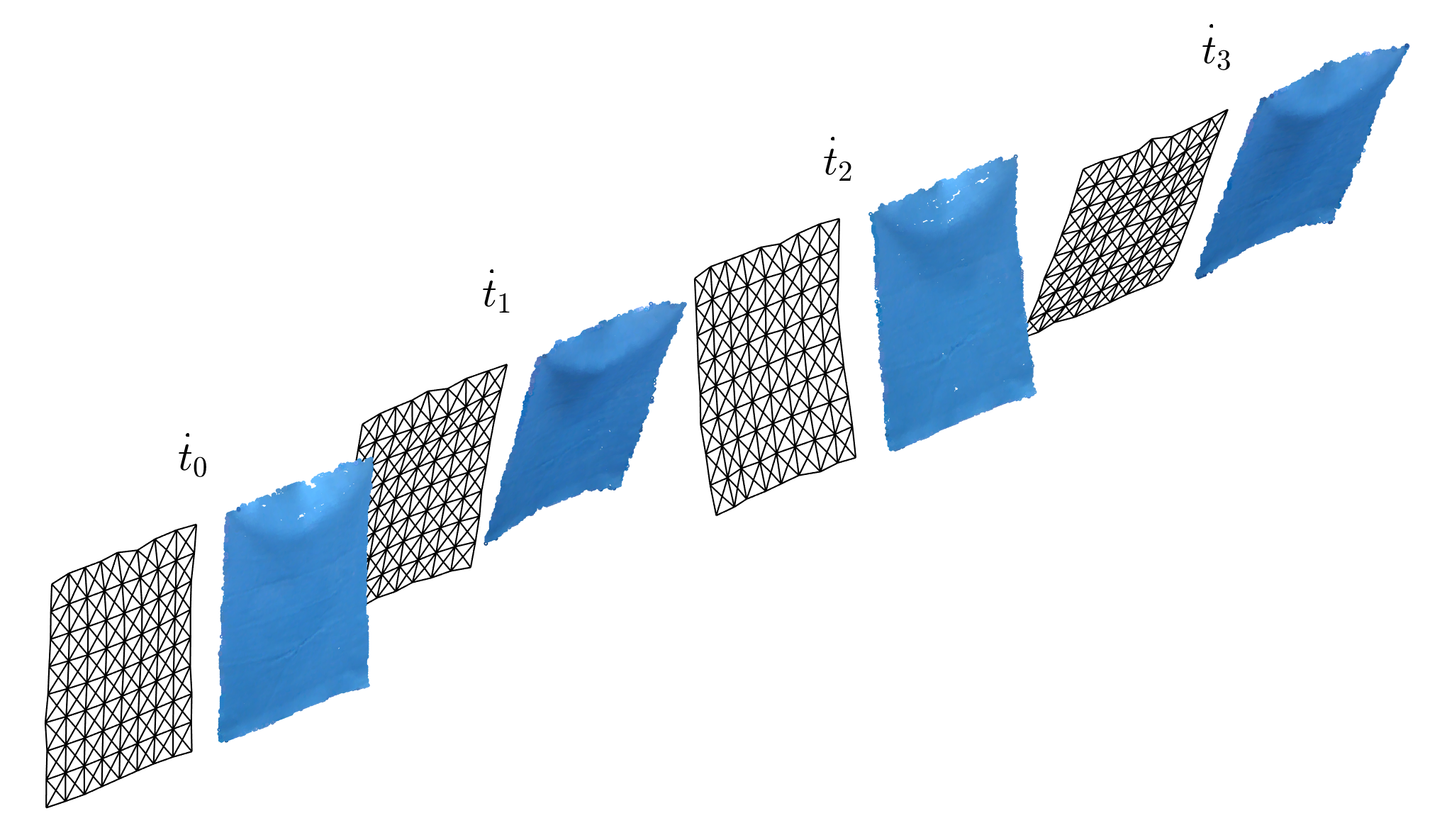}
	\caption{\label{cotton}Comparison at 4 time instants of the fast motion of light cotton (right) versus its simulation with the inextensible model (left) with $\delta = 0.52$ and $\alpha = 2.69$. The mean absolute error is $0.31$ cm.}
\end{figure}

As shown in Figure \ref{movimientos}, the error (\ref{error_rel}) for all four textiles and both motions are very low (its mean is under 5mm, see Table \ref{table:1}). For a visual comparison of the results, see Figure \ref{paper} and \ref{cotton}. A phenomenon that is noticeable is the concentration of error (clearly seen in the dispersion curves) at the beginning of the movement; likely due to the fact that air turbulences are more complicated at this stage. In regards to the parameter values, we note their discrepancy when comparing the fast and the slow motions (using the same textile), showing that they actually account for the aerodynamics of the motion. While this approach is not standard, note that using only $2$ parameters we are able to predict cloth dynamics accurately. 

It is relevant to emphasize that these results are obtained using a $9\times9$ mesh (with which our simulations are faster than real time) for a piece of cloth of size A3 ($29.7 \times 42\text{cm}^2$). Keeping the obtained optimal parameters, we also computed the error (\ref{error_rel}) for other three mesh resolutions: $9\times17$, $17\times9$ and $17\times17$, getting respectively: 0.37, 0.34 and 0.35cm as the mean of the absolute error (\ref{error_rel}) for the fast motion of light cotton (see Figure \ref{cotton}, with the $9\times9$ mesh the error is 0.31cm). This shows again (see Section \ref{section_cusick}) the robustness of the model with respect to mesh resolution. 

In the application of our model for control of robotic manipulation purposes, it may be necessary to simulate the motion of the garment prior to any possibility of calibration of its two relevant parameters. In such case, estimates of these parameters should be made from prior study of the model and of the involved fabric. This topic will be the subject of future work, but we have found that a simple, interim solution is already accurate to within our desired error bounds: as a prediction of motion regardless of cloth speed one may take as estimate of each parameter the arithmetic average of the values of its calibrations found in Table \ref{table:1} for a comparable fabric. Performing the simulations of both the slow and fast motions for light cotton using these a priori estimates instead of calibration, yielded $0.51$cm and $0.59$cm respectively as the mean of error (\ref{error_rel}) (compared to $0.41$cm and $0.31$cm with the optimal parameters). 

\section{Conclusions and further work}
In this work we have presented a continuous model for the simulation of inextensible textiles. Continuity is important because it guarantees that the model is very stable under different resolutions, allowing coarse meshes to be used (e.g. in robotic applications) soundly. On the other hand, inextensibility as implemented here permits the conservation of area within an error of less than $1\%$ without exhibiting \textit{locking}. We have also illustrated a framework for the empirical validation of this model. The results are accurate: even with a coarse $9\times9$ mesh the mean error stays under 5mm for different textiles and both fast and slow motions. Besides, the calibration depends only on two physical parameters. In a novel, and remarkably simple way, those two parameters take into account even the aerodynamic perturbations to motion, which for cloth are harder to model than for rigid, even vibrating, bodies. 

\smallskip

These properties make our model ideal for planning and control of cloth manipulation by robots. The authors intend to experiment further with varied types of cloth, fabrics, and motion, to learn about the meaning of the model parameters $\alpha, \delta$ and their relation to known factors that govern cloth behavior (such as the KES parameters) and aerodynamics (permeability, velocity of motion, geometry). This study should allow the use of the model as a simulator replacing the calibration of the parameters with an a priori forecast of their value at a very modest cost in terms of accuracy. We have already shown that with a crude forecast of their value our model already matches real motion within the $1$cm error margin.

\section{Acknowledgements}
This project was developed in the context of the project CLOTHILDE ("CLOTH manIpulation Learning from DEmonstrations") which has received funding from the European Research Council (ERC) under the European Union's Horizon 2020 research and innovation programme (grant agreement No. 741930) and is also supported by the Spanish State Research Agency through the María de Maeztu Seal of Excellence to IRI (MDM-2016-0656).

\bibliography{arxiv}   

\end{document}